\documentclass{article} % For LaTeX2e
\usepackage{iclr2025_conference,times}

\usepackage{amsmath,amsfonts,bm}

\def\eqref#1{equation~\ref{#1}}
\def\1{\bm{1}}

\DeclareMathAlphabet{\mathsfit}{\encodingdefault}{\sfdefault}{m}{sl}
\SetMathAlphabet{\mathsfit}{bold}{\encodingdefault}{\sfdefault}{bx}{n}

\usepackage{hyperref}
\usepackage{url}
\usepackage{tabularx}

\usepackage[utf8]{inputenc} % allow utf-8 input
\usepackage[T1]{fontenc}    % use 8-bit T1 fonts
\usepackage{hyperref}       % hyperlinks
\usepackage{url}            % simple URL typesetting
\usepackage{booktabs}       % professional-quality tables
\usepackage{amsfonts}       % blackboard math symbols
\usepackage{nicefrac}       % compact symbols for 1/2, etc.
\usepackage{microtype}      % microtypography
\usepackage{xcolor}         % colors

\usepackage{graphicx}
\usepackage{mathtools}
\usepackage{amssymb}
\usepackage{booktabs}
\usepackage{bbm}
\usepackage{wrapfig}
\usepackage{amsmath, bm}
\usepackage{algorithm, algpseudocode}
\usepackage{algorithmicx}
\usepackage{caption}
\usepackage{multirow}

\newcommand{\comment}[1]{}

\usepackage{calc}
\newlength\myheight
\newlength\mydepth

\title{GTR: Improving Large 3D Reconstruction Models through Geometry and Texture Refinement}

\author{
\centering
Peiye Zhuang
\And Songfang Han
\And Chaoyang Wang
\And Aliaksandr Siarohin 
\AND  Jiaxu Zou
\And Michael Vasilkovsky
\And Vladislav Shakhrai 
\And Sergei Korolev
\AND Sergey Tulyakov
\And Hsin-Ying Lee
}

\iclrfinalcopy 
\begin{document}

\maketitle

\vspace{-25pt}
\begin{center}
{\url{https://snap-research.github.io/GTR/}}
\end{center}
\vspace{-10pt}

\begin{figure}[h]
    \centering
    \includegraphics[width=\linewidth]{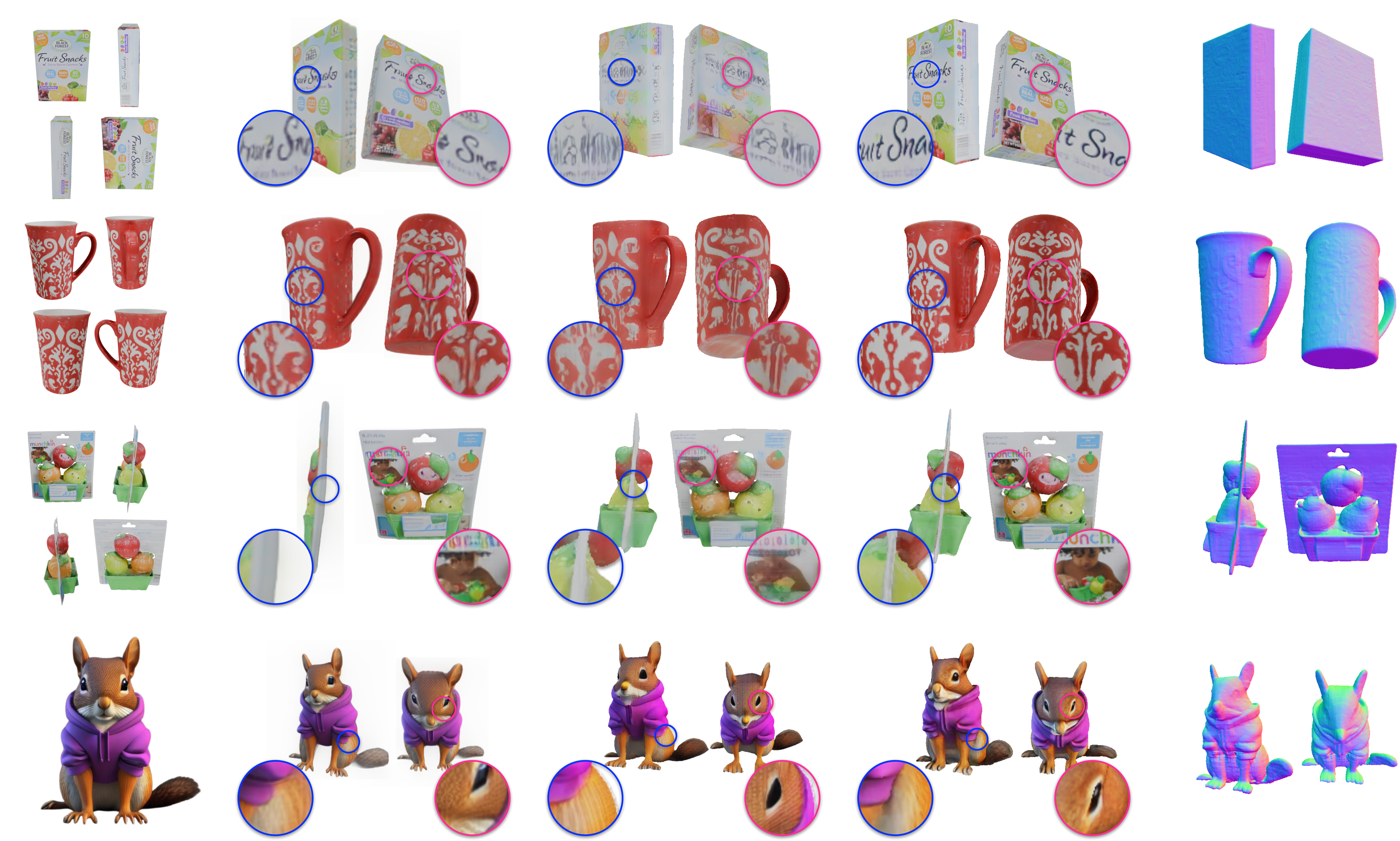}
    \hspace{.2cm} Input  \hspace{.1cm} 
     LRM~\citep{lrm}  \hspace{.1cm} 
     InstantMesh~\citep{instantmesh} \hspace{.2cm}   
     Ours \hspace{.8cm} Our normals
    \caption{\textbf{Comparison with the baseline methods on the sparse-view reconstruction task.} We visualize the novel view results generated from the GSO dataset~\citep{gso} (\textit{row 1-3}), or a generated image (\textit{row 4}). Note that for the generated image, we use Zero123++~\citep{zero123pp} to generate 6 views, which are used as input for InstantMesh~\citep{instantmesh} and GTR (Ours).
    }
    \label{fig:1}
\end{figure}

\begin{abstract}
\label{sec:abs}

We propose a novel approach for 3D mesh reconstruction from multi-view images. We improve upon the large reconstruction model LRM~\citep{lrm} that use a transformer-based triplane generator and a Neural Radiance Field (NeRF) model trained on multi-view images. We introduce three key components to significantly enhance the 3D reconstruction quality. First of all, we examine the original LRM architecture and find several shortcomings. Subsequently, we introduce respective modifications to the LRM architecture, which lead to improved multi-view image representation and more computationally efficient training.
Second, in order to improve geometry reconstruction and enable supervision at full image resolution, we extract meshes from the NeRF in a differentiable manner and fine-tune the NeRF model through mesh rendering. These modifications allow us to achieve state-of-the-art performance on both 2D and 3D evaluation metrics on Google Scanned Objects (GSO) dataset~\citep{gso} and OmniObject3D dataset~\citep{omniobject3d}.
Finally, we introduce a lightweight per-instance texture refinement procedure to better reconstruct complex textures, such as text and portraits on assets.
To address this, we introduce a lightweight per-instance texture refinement procedure. This procedure fine-tunes the triplane representation and the NeRF's color estimation model on the mesh surface using the input multi-view images in just 4 seconds. This refinement achieves faithful reconstruction of complex textures. Additionally, our approach enables various downstream applications, including text/image-to-3D generation.
\end{abstract}

\section{Introduction}
\label{sec:intro}

The task of generating 3D assets from text or images has wide applications in digital content generation and virtual reality (VR)~\citep{magic3d, inifinicity}.
Because of the scarcity of 3D data, efforts have been made to leverage pre-trained large-scale text-to-image diffusion models.
Some works learn 3D asset generation from the pre-trained image diffusion models via score distillation techniques~\citep{dreamfusion,prolificdreamer,sjc,fantasia3d,magic3d,hifa,magic123}, which, however, require from several minutes to several hours for a single asset, since the optimization algorithm requires many iterations.
Other works propose to fine-tune image diffusion models into a multi-view image diffusion model~\citep{zero123, mvdream, syncdreamer,wonder3d} using 3D asset datasets~\citep{objaverse}.
These multi-view images serve as an intermediate 3D representation, which is taken as input by large 3D reconstruction models (LRMs)~\citep{lrm,instant3d,dmv3d,crm,lgm} for asset generation. However, these previous approaches struggle to reconstruct faithful textures and high-quality geometry when using the Marching Cube (MC) algorithm~\citep{mc}. Moreover, after extracting meshes using MC, the texture quality degrades even further. 

In contrast, we start developing the feed-forward mesh generation model by carefully examining the standard LRM architecture. 
First, we observed that DINO features tend to discard high-frequency image details, which are important for the precise reconstruction, from the input images. 
Thus we replace the pre-trained DINO transformer~\citep{DINO} used in previous LRMs~\citep{lrm,instant3d,dmv3d,crm,lgm} with a convolutional encoder for the multi-view images. 
Moreover, because of the high computation requirements of the transformers, previous methods usually run a transformer at $32^2$ triplane resolution and use a deconvolution to upsample this triplane. However, we noticed that reconstruction from standard LRM often exhibits regular grid artifacts (see Fig.~\ref{fig:1}). We speculate that the nature of these artifacts is similar to grid-shape artifacts observed in 2D deconvolutional generators~\citep{deconv}. 
To address this, we replace all deconvolution layers with Pixelshuffle layers~\citep{pixelshuffle}. Finally, we employ two shallow Multi-layer Perceptrons (MLPs) to separately predict density and colors, which is beneficial for our following fine-tuning stages, which we will explain shortly. 

Learning meshes is significantly harder than learning a NeRF. Thus, we begin by training this modified architecture using NeRF volume rendering. With the trained NeRF model in place, we proceed to fine-tune the pipeline using mesh rendering (also known as rasterization). To achieve this, we employ Differentiable Marching Cubes (DiffMC)~\citep{diffmc} to extract meshes from the NeRF density fields by transferring the densities to a signed distance function (SDF) representation. This enables us to render full-resolution images for supervision. Additionally, we employ a depth loss to guide the geometry extraction. 
Our feed-forward mesh generation pipeline significantly boosts the quality of the reconstructions compared to the results extracted from NeRFs using MC.

While the feed-forward mesh generation pipeline achieves advanced 3D reconstruction quality, it still faces challenges in accurately reconstructing intricate textures, such as text and complex patterns. 
To address this limitation and further enhance texture quality, we introduce a straightforward yet highly effective texture refinement procedure. Specifically, we fine-tune both the triplane feature and the color estimation model for each instance using sparse multi-view input images. 
As mentioned before, the shallow and separate density and color estimation models enable efficient updating of colors for surface points on the extracted meshes. In other words, the heavy triplane generator remains fixed in this texture refinement stage. This enables us to achieve rapid optimization, reaching 5 iterations per second on an A100 GPU.
Remarkably, our method achieves faithful texture reconstruction with just $20$ steps of fine-tuning on 4-view images, requiring a mere 4 seconds on an A100 GPU.

Our proposed approach enables faithful 3D reconstruction from the multi-view input images, as shown in Fig.~\ref{fig:1}.
We conduct a comprehensive comparison of our method with multiple concurrent works~\citep{lrm, instantmesh, lgm} using the Google Scanned Object (GSO)~\citep{gso} and the OmniObject3D~\citep{omniobject3d} datasets, employing various evaluation metrics. For instance, in the 4-view reconstruction task using GSO dataset~\citep{gso},
our approach achieves a Peak Signal-to-Noise Ratio (PSNR) of 29.79, a Structural Similarity Index (SSIM) of 0.94 and a Learned Perceptual Image Patch Similarity (LPIPS) score of 0.059.

% including Peak signal-to-noise ratio (PSNR), Structural Similarity Index (SSIM), Learned Perceptual Image Patch Similarities (LPIPS) in 2D image space, and Chamfer Distance (CD) and Intersection over Union (IoU) in 3D geometry space. 
Extensive experiments show that our feed-forward model achieves superior results compared to the baseline approaches, while our per-instance refinement approach enables further texture improvement on text and complex patterns. See Fig.~\ref{fig:1} for visual examples. 
% We show ${18\%}$ improvement in PSNR, ${30\%}$ improvement in LPIPS, and ${33\%}$ better score in CD comparing to previous state-of-the-art.
%\sergey{Do we want to add any comparisons here? We show 25\% better scores than X...?}. 

We summarize our technical contributions to two key components of multiview-to-3D reconstruction: (1) mesh generation and (2) accurate texture reconstruction. These contributions are outlined as follows:

\begin{itemize}
    \item We introduce a holistic design for multi-view image to 3D reconstruction to enhance the quality of the generated meshes. This includes modifications to the existing LRM model and fine-tuning the NeRF model with a differentiable mesh representation.
    \item We present an efficient per-instance texture refinement process, leveraging input images to enhance texture details.
\end{itemize}

Furthermore, our model can be adapted to various downstream applications, such as text/image-to-3D generation tasks.

\begin{figure}[t]
    \centering
    \includegraphics[width=\linewidth]{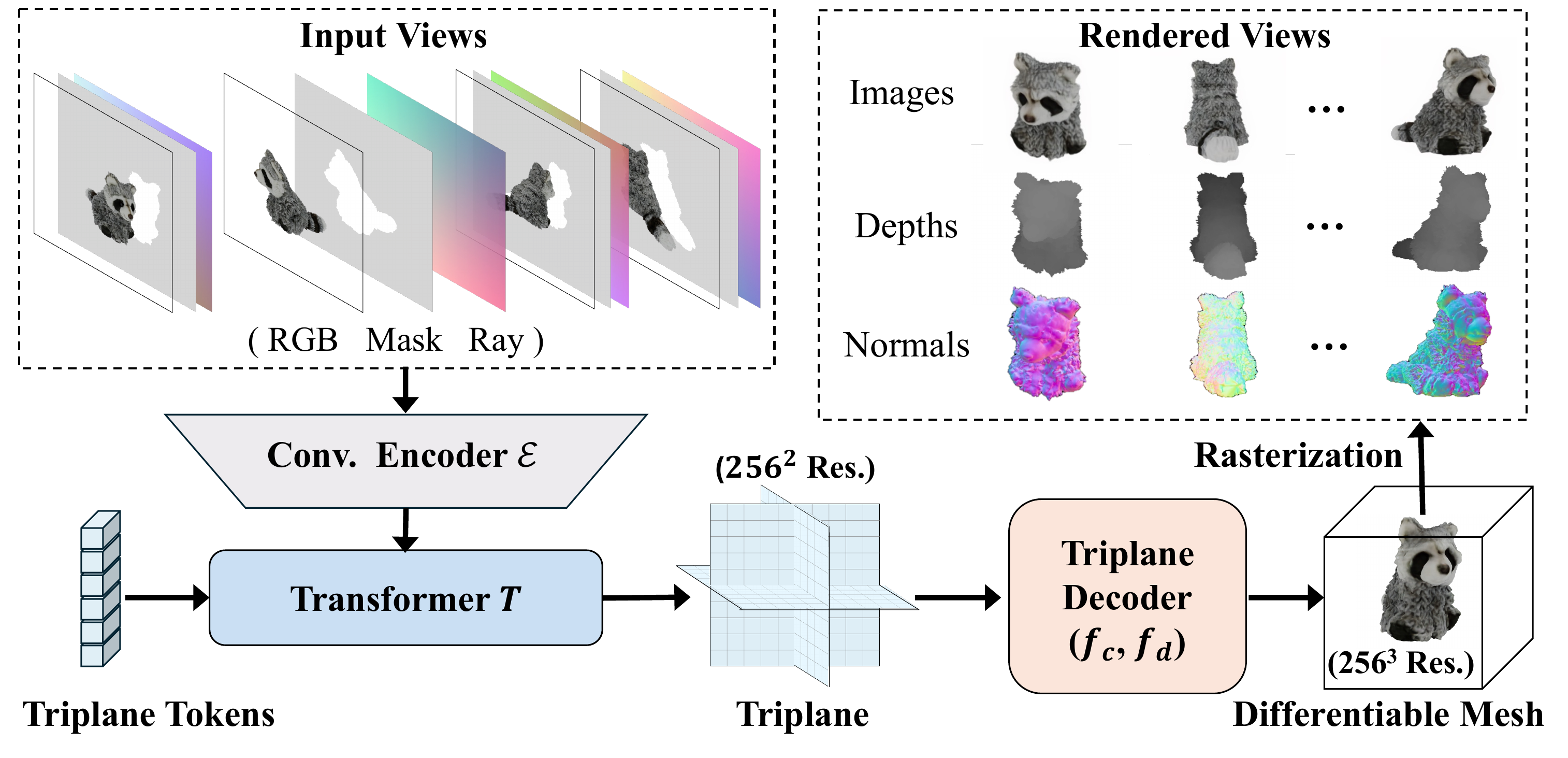}
    \caption{\textbf{Overview of our proposed approach for 3D reconstruction.} Our pipeline consists of a convolutional decoder $\mathcal{E}$, a transformer-based triplane generator, $T$, and a NeRF-based triplane decoder that contains two MLPs, $f_c$ and $f_d$, for color and density prediction, respectively. In practice, the triplane resolution is set to $256$, and the mesh representation has a grid size resolution of $256$.}
    \label{fig:2}
\end{figure}

 \vspace{-.3cm}
\section{Related Work}
\label{sec:related}
 \vspace{-.3cm}

\textbf{Optimization-based 3D generation} aims to use pre-trained large-scale text-to-image diffusion models~\cite{ldm,imagen} for 3D generation, given the insufficient scale and diversity of existing 3D datasets.
To distill 3D knowledge from the text-to-image models, a Score Distillation Sampling (SDS) approach and its variants~\cite{dreamfusion,sjc,prolificdreamer,fantasia3d,hifa} have been proposed. In these methods, noise is added to an image rendered from 3D models like NeRFs and subsequently denoised by a pre-trained text-to-image generative model~\cite{ldm}. The SDS approach aims to minimize the Kullback-Leibler (KL) divergence between a prior noise distribution and the estimated noise distribution from the text-to-image model. However, these SDS-based methods are time-consuming, usually taking up to hours to generate a single instance. Alternatively, feed-forward 3D generation models have been proposed to achieve faster generation.

\textbf{Feed-forward 3D generation} has gained increasing attention recently due to its speed advantage. 
Most existing feed-forward 3D generation pipelines consist of two stages: (i) a text prompt (or a single image) to multi-view generation and (ii) multi-view images to 3D shape reconstruction. In the first stage, multi-view images are generated from text or image input using a multi-view generator~\cite{zero123,mvdream,syncdreamer, wonder3d}, which are usually fine-tuned from image or video diffusion generation models~\cite{ldm, svd}.
In the second stage, a 3D instance is reconstructed from input multi-view images.
In this case, various 3D representations are used such as neural implicit fields~\cite{instant3d,dmv3d,lrm}, Gaussian Splatting~\cite{grm, lgm}, or meshes~\cite{crm, instantmesh}.
However, existing methods encounter challenges to either faithfully reconstruct textures when using implicit field representations~\cite{instant3d,dmv3d,lrm}, or difficult to extract explicit geometries when using Gaussian Spatting~\cite{grm, lgm}. 
In this work, we focus on the second stage of the pipeline, i.e., reconstructing 3D shapes from multi-view images. To address those challenges, we propose a novel 3D reconstruction approach that improves the 3D quality through geometry and texture refinement. 

\textbf{Differentiable mesh} is a hybrid 3D representation that combines implicit and explicit surface representations, i.e., SDFs and meshes, and is suitable for 3D optimization. Recent popular representations include DMTet, Flexicubes, Differentiable Marching Cubes (DiffMC)~\cite{dmtet,flexicube,diffmc}. In our work, we chose DiffMC~\cite{diffmc} as it doesn't require additional components beyond our existing model pipeline, in contrast to using DMTet and Flexicubes~\cite{dmtet,flexicube}, where a deformation and a weight prediction net are required.

\textbf{MVS-based 3D reconstruction} aims to generate novel views from sparse-view input images using Multi-view Stereo (MVS) techniques.
Classic MVS methods leverage cost volumes~\cite{kutulakos2000theory,seitz1999photorealistic}, point clouds~\cite{stereopsis2010accurate,lhuillier2005quasi}, or depth maps~\cite{campbell2008using,gallup2007real} to learn blending weights for input sparse-view pixels for generating novel views.
Recently, learning-based methods~\cite{gu2020cascade,ma2021epp,wang2021patchmatchnet,wei2021aa,yi2020pyramid} have been proposed that can generalize to novel scenes.
However, these methods require input views to have dense local overlap and struggle to generate 360-degree views of 3D assets.
It is even more challenging when input multi-view images are generated without precise pixel-level alignment. Alternatively, in our work, we propose a simple yet effective texture refinement procedure that enables high-quality texture reconstruction from sparse-view input and is robust to synthetic images.

\begin{figure}[t]
    \centering
    \includegraphics[width=\linewidth]{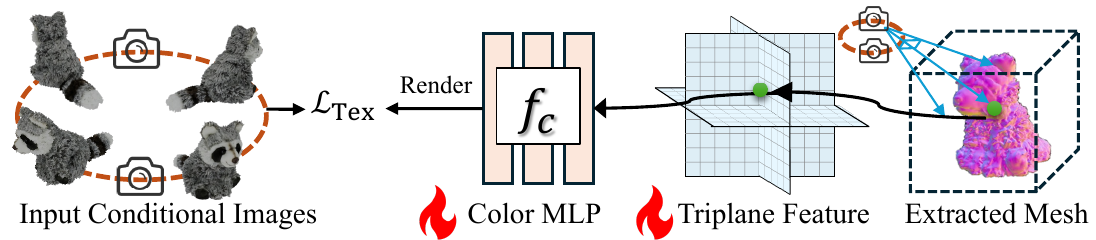}
    \caption{\textbf{Texture refinement for extracted meshes.} We refine the texture by fine-tuning the triplane feature of the asset and the color MLP, $f_c$, using the input images. We use an L2 loss on the images, defined as $\mathcal{L}_\text{Tex}$. }
    \label{fig:3}
\end{figure}

%
 % \vspace{-.3cm}
\section{Method}
\label{sec:method}
 % \vspace{-.3cm}

We separate the technical details of our method into three parts. In Sec.~\ref{sec:stage1}, we explain the modifications made to the existing LRM architecture. In Sec.~\ref{sec:stage2}, we present a training procedure for our feed-forward mesh generation model. The overview of our feed-forward mesh generation pipeline is illustrated in Fig.~\ref{fig:2}.  Finally, in Sec.~\ref{sec:stage3}, we introduce our per-instance texture refinement procedure. This procedure is highlighted in Fig.~\ref{fig:3}.

\subsection{Improving the large 3D reconstruction architecture}
\label{sec:stage1}

A standard LRM~\citep{lrm} architecture consists of an image encoder, a transformer-based triplane generator with a deconvolutional triplane upsampler, and a NeRF-based triplane decoder. In our work, we propose several modifications to the LRM architecture outlined next.

\textbf{Convolutional image encoder} $\mathcal{E}$. LRM~\citep{lrm, instant3d, dmv3d} models typically utilize pre-trained transformer network DiNO ViT~\citep{DINO}. However, we observe that since DiNO ViT was designed for semantic understanding, it tends to ignore some local details irrelevant to image semantics but required for accurate reconstruction. Thus we propose to replace the DiNO~\citep{DINO} architecture with a convolutional encoder. Since this encoder is trained from scratch along with other components it does not exhibit the bias of pre-trained DiNO. An additional advantage of this encoder is that it does not require any modifications to consume additional inputs useful for 3D reconstruction. To this end, we complement the input images with binary foreground mask and Plücker coordinates for camera rays. We show a comparison of the training process using different encoders in Appendix~\ref{sup:ablation}.

\textbf{Triplane upsampler} $T$. One of the LRM architecture shortcomings is the tendency to generate grid-shaped artifacts. We attribute this issue to the deconvolution operation utilized in a triplane upsampler. Indeed, to reduce the computational requirements, the original LRM proposed to run a transformer-based triplane generator on $32^2$ resolution and later utilize deconvolution operation to upsample the triplane. The deconvolution operation is widely studied in GAN literature, for example, Odena et al.~\citep{deconv} show that for 2D generators deconvolutions are the main source of grid-shaped artifacts. To this end, we replace the deconvolution upsampling with a linear layer followed by a pixelshuffle~\citep{pixelshuffle}. This simple modification helps to alleviate grid-shaped texture artifacts. 
%See the comparison in  Fig.~\ref{fig:1}.

\begin{figure}[t]
    \centering
    \includegraphics[width=\linewidth]{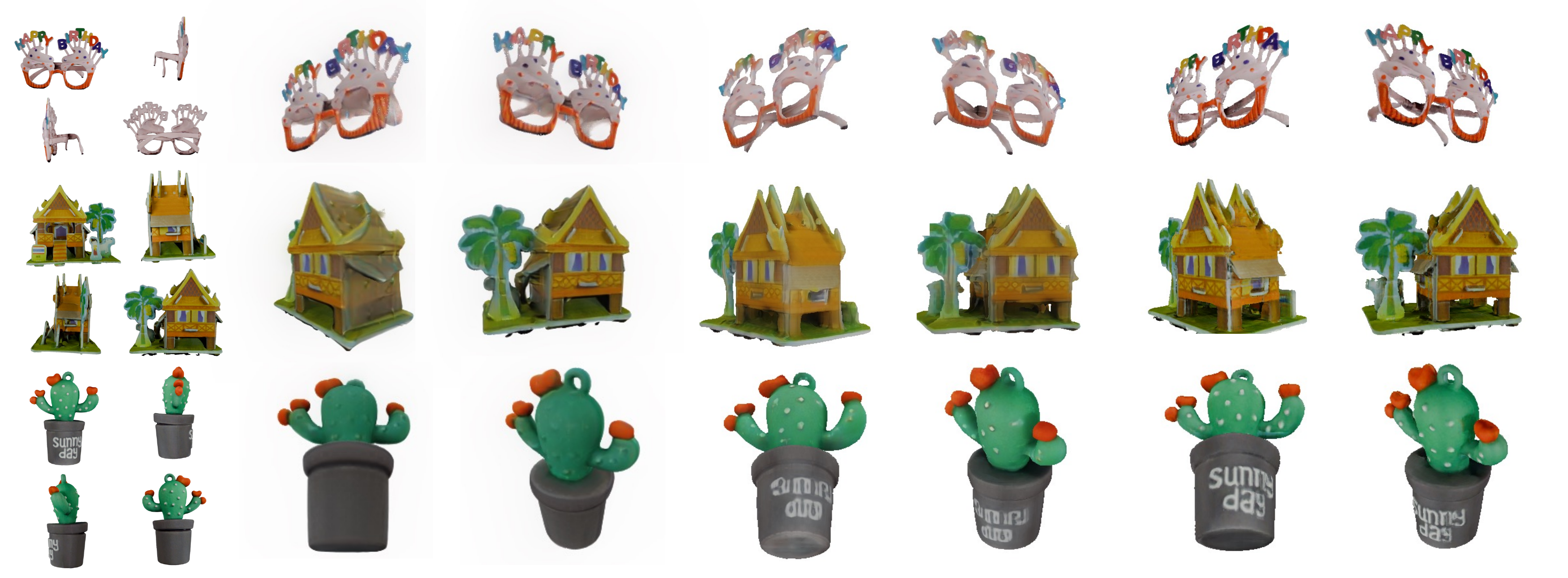}
     \hspace{1cm} Input  \hspace{1.cm} 
     LRM~\citep{lrm}  \hspace{.5cm}  
     InstantMesh~\citep{instantmesh}  \hspace{.5cm}    
     Ours \hspace{.5cm}
    \caption{\textbf{Comparison with the baseline methods on OmniObject3D dataset}~\citep{omniobject3d}. We compare the novel view reconstruction results with the baseline methods. LRM~\citep{lrm} takes the front-view image as input while Instantmesh~\citep{instantmesh} and our method take 4 views.}
    \label{fig:omni}
\end{figure}

 % \caption{\textbf{Comparison with the baseline methods on the sparse-view reconstruction task.} We visualize the novel view results generated using either ground-truth 4-view images (\textit{row 1-3}) from the GSO dataset~\citep{gso}, or a single generated image (\textit{row 4}).

% \textbf{NeRF decoders} $\{f_c, f_d\}$. Differently from LRM, we employ two separate MLPs, defined as $f_c$ and $f_d$, to estimate colors and density, respectively. This  change does not affect the performance significantly, it has a more utilitarian purpose, we can train $f_c$ and freeze $f_d$ if we want to finetune the texture, or other way around if we want to finetune the geometry.

\textbf{NeRF decoders} ${f_c, f_d}$. Unlike previous LRMs~\citep{lrm, instant3d}, we utilize two separate MLPs, defined as $f_c$ and $f_d$, to estimate colors and density, respectively. This modification does not impact performance; however, it serves a more practical purpose. For instance, we can train the color model $f_c$ and freeze $f_d$ when fine-tuning asset texture, or vice versa if fine-tuning asset geometry.

\subsection{Feed forward mesh generation model}
\label{sec:stage2}
The optimization through mesh representation may pose a significant challenge. Indeed, the gradients for backpropagation through mesh exist only in a small local neighborhood and, thus, convergence heavily depends on accurate initialization. 
% To this end
To tackle this, we develop a two-stage training procedure, where in the first stage we utilize the volumetric rendering and optimize NeRF, and, in the second stage, we perform geometry refinement by optimizing through a mesh representation.

\textbf{NeRF training stage}. In this stage, we simply optimize our modified architecture (see Sec.~\ref{sec:stage1}) using a mean squared error (MSE) loss and an LPIPS loss on images, defined as
\begin{equation}
    \mathcal{L_\text{NeRF}}  = \quad \mathcal{L}_\text{rgb} + \lambda_\text{p} \ \mathcal{L}_\text{LPIPS},
\end{equation}
where $\lambda_\text{p}$ denotes the loss weight.

% We then present the end-to-end geometry refinement, initialized using the learned NeRFs, in the following.

\textbf{Geometry refinement with NeRF initialization}.
In the geometry refinement stage, we fine-tune the entire pipeline using mesh rendering. Specifically, we transfer the density field to an SDF field:
\begin{equation}
    \text{sdf} =\quad  - (d - s),
\end{equation}
    where $d$ is the estimated density and $s$ is a pre-defined level set for DiffMC, $s=10$, in practice.

Next, we render images, depths, and masks for training via mesh rendering (a.k.a rasterization). 
We fine-tune the pipeline using an MSE loss and an LPIPS loss on images, MSE losses on depths, masks, and normal maps.
Formally, we define the loss as:

% a regularization loss on opacity~\citep{meshlrm}
\begin{equation}
    \mathcal{L_\text{Mesh}} = \quad \mathcal{L}_\text{rgb} + \lambda_p \ \mathcal{L}_\text{LPIPS} \ 
    + \lambda_d \ \mathcal{L}_\text{depth} \
    + \lambda_m \ \mathcal{L}_\text{mask} + 
    + \lambda_n \ \mathcal{L}_\text{normal} 
\end{equation}
where $\lambda_\text{p}$, $\lambda_\text{d}$, $\lambda_\text{m}$ and $\lambda_n$ are the loss weights.

% \\lambda_r \ \mathcal{L}_\text{reg},

\subsection{Texture refinement for mesh representations}
\label{sec:stage3}

\begin{figure}[t]
    \centering
    \includegraphics[width=\linewidth]{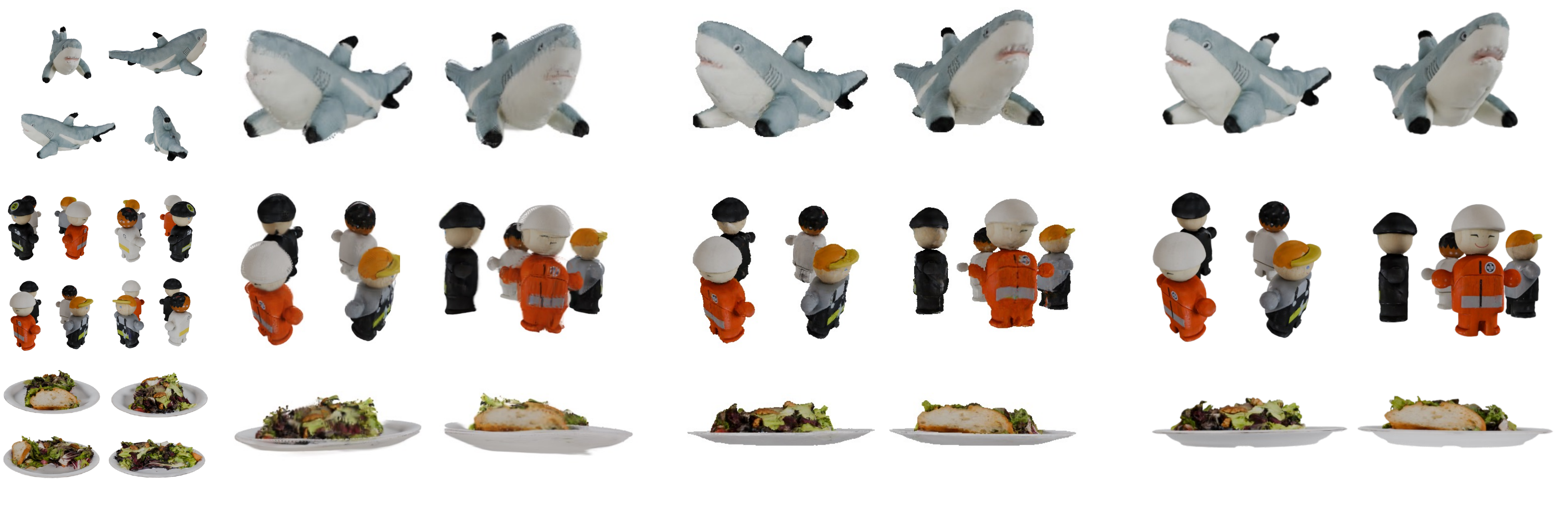}
    Input \hspace{.9cm} 
    LGM~\citep{lgm} \hspace{2 cm}
    Ours \hspace{3 cm} 
    GT  \hspace{1.5 cm}
    \caption{\textbf{Comparison with LGM~\citep{lgm} on GSO~\citep{gso} dataset}. Both LGM~\citep{lgm} and our method take 4-view images as input and reconstruct novel views.}
    \label{fig:lgm}
\end{figure}

% While the geometry refinement stage enables supervision at the full image resolution and the SD encoder allows global image representations, we observe that such an end-to-end model still struggles with faithful texture reconstruction, particularly when the textures contain complex and high-frequency patterns. To address this,

Inspired by previous works that utilize the Gaussian Splatting~\citep{lgm, grm} representation, where colors of the input images can be easily retained in Gaussian features, we notice a disparity in our pipeline, which lacks this color memorization scheme. Thus, we refine the triplane feature of an asset and the color MLP, $f_c$, using the input multi-view images, $\bm{I}_\text{cond}$, for surface points on the extracted mesh. We illustrate the refinement procedure in Fig.~\ref{fig:3}.
The separated density and color MLPs benefit the texture refinement procedure, as we only fine-tune a single MLP. We use an MSE loss on input images for texture refinement:

\begin{equation}
\mathcal{L}_\text{Tex} = \quad \mathcal{L}_2 (\bm{I}_\text{cond}, \ \bm{\hat{I}_\text{cond}}),
\end{equation}
where $\bm{I}_\text{cond}$ and $\bm{\hat{I}_\text{cond}}$ denote the ground-truth and predicted input images, respectively. Note that at this stage, the image encoder $\mathcal{E}$ and the triplane generator $T$ are fixed and used to generate the initial triplane features of assets. The density estimation MLP, $f_d$, is also fixed and used to extract meshes.

\section{Experiments}
\label{sec:exp}

In Sec.~\ref{sec:41}, we explain training datasets, implementation details, and evaluation with the baseline methods.
We present both quantitative and qualitative results in Sec.~\ref{sec:42} and the ablation study in Sec.~\ref{sec:43}. Additional implementation details and comparisons are presented in Appendix.

% Finally, we show additional downstream applications using our method in Sec.~\ref{sec:44}

\subsection{Experiment Settings}
\label{sec:41}

\textbf{Dataset}: Our model is trained on a 140k asset dataset, which merges the filtered Objaverse dataset~\citep{objaverse} with an internal 3D asset dataset. We filter the high-quality Objaverse dataset to retain 26k superior assets of high quality. We render 32 random views for each training asset. In Appendix~\ref{sup:ablation}, we also provide additional ablation studies with our model trained solely on the Objaverse dataset containing 100k assets.

\textbf{Implementation Details}. In practice, the loss weights are set to $\lambda_{p}=0.5, \lambda_d=0.5, \lambda_m=1 \text{ and } \lambda_n = 1$. In practice, we found that adding the normal loss can lead to unstable training. Therefore, during the geometry refinement stage, we add the normal loss once the model has stabilized, and we freeze the generator when incorporating the normal loss. Input multi-view images are of $512$ resolution. 
The triplane transformer contains 24  attention blocks with a hidden dimension of $1024$. 
Each attention layer has 16 attention heads and each head has a dimension of 64. During the NeRF training stage, images are rendered at $512$ resolution, and the NeRF model is trained using a patch size of $128^2$. We uniformly sample $256$ points along each camera ray. The density and color MLPs consist of 3 and 4 layers, respectively, with a hidden size of 512. 

In the NeRF training stage, we use an AdamW optimizer with a learning rate $1e-4$ and a weight decay of 0.05. Cosine scheduling is employed to gradually reduce the learning rate to 0 after 150k training iterations. 
We use a batch size of 512 on 32 A100 GPUs.
For each asset, we randomly choose 4 views as input and another 4 views for supervision.
In the geometry refinement stage, we choose a grid size of 256 during mesh extraction using DiffMC. We use another AdamW optimizer with a learning rate $5e-5$. The batch size is 192 on 32 A100 GPUs. For each asset, we randomly choose 4 views as input and another 8 views for supervision.
In the per-instance texture refinement stage, the learning rates for the triplane feature and the color MLP, $f_c$, are $0.15$ and $1e-4$, respectively.

\begin{table}[t]
    \centering
    \caption{\textbf{Quantitative comparison on GSO dataset}~\citep{gso}.}
    % \vspace{-.3cm}
       \label{tab:gso}
    \setlength{\tabcolsep}{3pt}
    \begin{tabular}{l c c c c | c c}
    \midrule
    Method && PSNR ($\uparrow$) & SSIM ($\uparrow$) & LPIPS ($\downarrow$) & CD ($\downarrow$) & IoU ($\uparrow$) \\ \midrule
        LRM~\citep{lrm} && 20.446 & 0.904 &0.126 & 6.383 & 0.352\\
        % SV3D~\citep{sv3d} && 22.098 & 0.898 & 0.119 & 0.177 & 0.882 \\ 
        SV3D~\citep{sv3d} && 22.098 & 0.898 & 0.119 & 1.770 & 0.682 \\ 
        \midrule
        % CRM~\citep{crm} && 22.195 & 0.891 & 0.150 & 0.252 & 0.787 \\ 
        CRM~\citep{crm} && 22.195 & 0.891 & 0.150 & 1.252 & 0.617 \\ 
        LGM~\citep{lgm} && 25.227 & 0.925 & 0.066 &1.373  &0.601 \\ 
        InstantMesh (NeRF) ~\citep{instantmesh}&&24.740 &0.923 &0.080 &1.101 &0.635 \\ 
        InstantMesh (Mesh)~\citep{instantmesh} && 24.444 & 0.920 & 0.08 & 1.115 & 0.645 \\ 
        \midrule
        Ours (Feed-forward) && 28.673 & 0.946 &0.055 
        &\multirow{2}{*}{\textbf{0.740}} & \multirow{2}{*}{\textbf{0.708}}\\ 
        % & 0.7404 &0.6517 \\
        Ours (Tex. refine) &&\textbf{29.788} &\textbf{0.960} & \textbf{0.047} & &\\
        % &\textbf{0.7404} &\textbf{0.6517} \\
        \midrule
    \end{tabular}
\end{table}

\textbf{Evaluation}. We evaluate our method alongside baseline methods, including LRM~\citep{lrm}, SVD~\citep{sv3d}, CRM~\citep{crm}, InstantMesh~\citep{instantmesh}, and LGM~\citep{lgm} using the Google Scanned Objects (GSO) ~\citep{gso} and OmniObject3D~\citep{omniobject3d} dataset. We use the identical data lists of the GSO and OmniObject3D dataset and render camera orbits as outlined in Instantmesh~\citep{instantmesh}. Specifically, 300 GSO assets and 130 OmniObject3D assets (from 30 classes) are used for evaluation. We render 20 images for each asset in a trigonometric orbiting trajectory, i.e., maintaining uniform azimuths and elevations in $\{-30^{\circ}, 0^{\circ}, 30^{\circ}\}$. We use PSNR, SSIM, and LPIPS as image evaluation metrics, while Chamfer Distance (CD) and mIoU are utilized for 3D geometry evaluation. 
To evaluate 3D geometry, we follow the mesh processing steps in InstantMesh~\citep{instantmesh}. Specifically, we reposition the generated meshes to the origin and align the coordinate system with the ground-truth meshes. We then rescale all meshes into a $[-1, 1]^3$ cube. We also use Iterative Closest Point (ICP) registration to align the generated meshes to the ground truth meshes.
We sample 16000 points on the asset surface to compute the CD and IoU scores.

\vspace{-.3cm}
\subsection{Results}
\label{sec:42}
\vspace{-.3cm}

\begin{figure}
    \centering
    \vspace{-.3cm}
    \includegraphics[width=\linewidth]{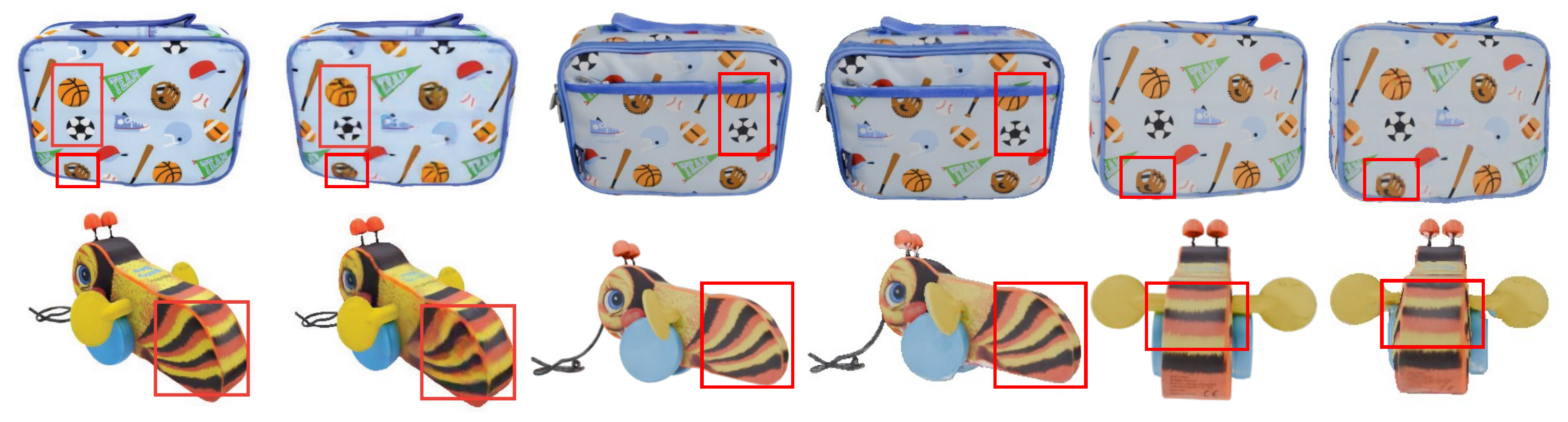}
    \centering
   \hspace{1cm} GT\hspace{1.7cm} MeshLRM 
    \hspace{1.3cm} GT \hspace{1.5cm}  Ours
    \hspace{1.3cm} GT \hspace{1.5cm}  Ours
    \caption{\textbf{Visual comparison with MeshLRM}~\citep{meshlrm} on GSO dataset~\citep{gso}. Our method generates better (at least comparable) textures with more details than the concurrent work~\citep{meshlrm}.}
    \label{fig:meshlrm}
\end{figure}

\textbf{Qualitative evaluation}. We compare our method with baseline approaches using the GSO~\citep{gso} and OmniObject3D~\citep{omniobject3d} datasets in Fig.\ref{fig:1}, Fig.\ref{fig:omni}-\ref{fig:lgm}. We observe that our method achieves more faithful texture reconstruction with finer details and more accurate geometry. For instance, in Fig.~\ref{fig:1}, our model can generate clear text on the first example and complex texture patterns on the second example. In the third example, our model enables the reconstruction of portraits printed on the asset. 
In Fig.\ref{fig:lgm}, we compare our method with LGM~\citep{lgm}, which uses Gaussian Splatting as a 3D representation. We observe that while LGM~\citep{lgm} can generate high-quality textures, it often tends to generate floating Gaussian points in inaccurate regions, even when the input images are ground-truth multi-view images. In Fig.~\ref{fig:meshlrm}, we present a comparison between our results and those of the concurrent MeshLRM approach\citep{meshlrm}. The visual results indicate that our methods produces outcomes comparable to those achieved by MeshLRM.
We present additional visual results in Appendix~\ref{sup:visual}.

\textbf{Quantitative evaluation}. We present the evaluation scores for the GSO~\citep{gso} and OmniObject3D~\citep{omniobject3d} dataset in Tab.~\ref{tab:gso} and Tab.~\ref{tab:omni}, respectively. 
The CD scores are presented by multiplying a rescale factor of $100$. Note that both  LGM~\citep{lgm} and InstantMesh~\citep{instantmesh} baselines are concurrent works. 
For InstantMesh~\citep{instantmesh}, we compare both their results using either neural rendering or mesh rendering.
Since LRM~\citep{lrm} takes a single image as input, we provide a front-view image to it.
Our full approach achieves the best evaluation results in 2D and 3D evaluations. We also present the evaluation scores using our feed-forward model, i.e., without the texture refinement procedure. These results still show significant improvements in texture and geometry quality. 

Additionally, the LGM~\citep{lgm} method serves as a strong baseline as we directly compare against their generated images rather than rendered images from extracted meshes. We note that the baselines that do not incorporate some explicit geometry representation during training typically yield inferior rendering results when extracted to meshes.  
However, our method achieves better results than their directly generated images. 
Furthermore, it takes 1 minute to extract meshes using LGM~\citep{lgm} from Gaussian Splatting. In contrast, our method enables to generate meshes within 1 second, plus an additional 4 seconds for texture refinement, which in total is still faster than LGM~\citep{lgm} inference.

% \subsection{Applications}
% \label{sec:44}

\textbf{Applications}. Our method enables downstream tasks such as text/image-to-3D generation. In this case, we generate multi-view images using pre-trained text-to-image and/or image-to-multiview diffusion models~\citep{ldm, zero123pp}. In practice, we use Zero123++~\citep{zero123pp} to generate 6-view images as the input for our model. We show the generated results in  Fig.~\ref{fig:app} and additional results in the Appendix~\ref{sup:visual}.

\begin{table}[t]
    \centering
    \caption{\textbf{Quantitative comparison on OmniObject3D dataset}~\citep{omniobject3d}.}
    % \vspace{-.3cm}
       \label{tab:omni}
    \setlength{\tabcolsep}{3pt}
    \begin{tabular}{l c c c c | c c}
    \midrule
    Method && PSNR ($\uparrow$) & SSIM ($\uparrow$) & LPIPS ($\downarrow$) & CD ($\downarrow$) & IoU ($\uparrow$) \\ \midrule
        LRM~\citep{lrm} &&18.082 &0.888 &0.125 & 4.347 & 0.448\\
        \midrule
        LGM (GS)~\citep{lgm}  && 22.826 & 0.913 & 0.068 &0.893 & 0.626 \\ 
        InstantMesh (NeRF) ~\citep{instantmesh}  && 22.609 &0.914 &0.076 &0.660 & 0.671  \\ 
        InstantMesh (Mesh)  ~\citep{instantmesh} && 22.141 &0.910 &0.079 & 0.603 & 0.675  \\ 
        \midrule
        % Ours (NeRF vanilla)  && 26.2985 & 0.9361 & 0.0576 & 0.5042 & 0.7279  \\
        Ours (Feed-forward)  && 25.372 & 0.931 & 0.06
        &\multirow{2}{*}{\textbf{0.504}} & \multirow{2}{*}{\textbf{0.728}} \\
        Ours (Tex. refine)  && \textbf{25.400} &\textbf{0.937} &\textbf{0.059} & &   \\
        \midrule
    \end{tabular}
 \vspace{-.3cm}
\end{table}

\begin{figure}[t]
    \centering
    \includegraphics[width=\linewidth]{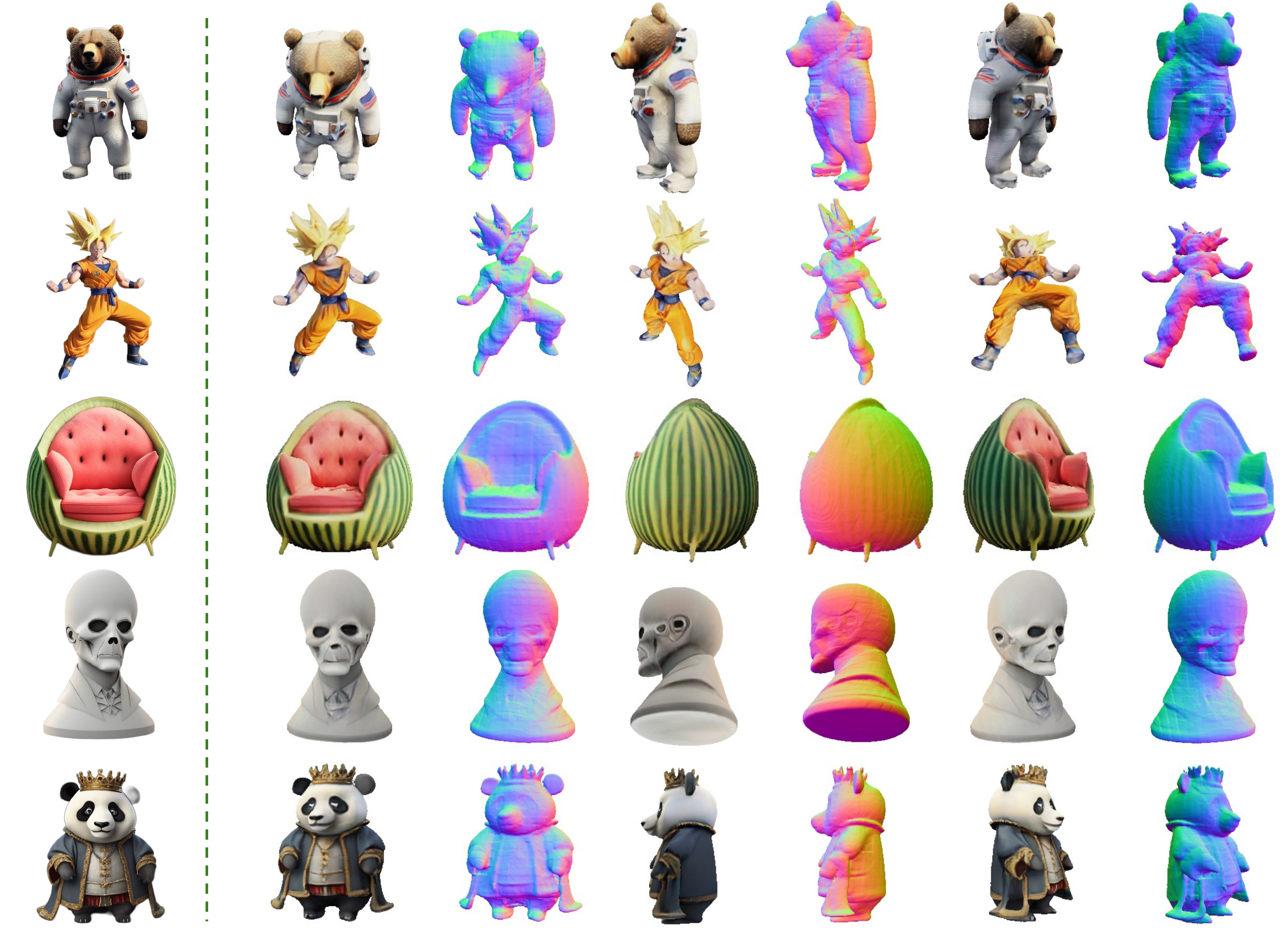}
    Input \hspace{4cm} Synthetic novel views \hspace{4cm}
    
\caption{\textbf{Image-to-3D generation}. Our method can adapt to the text/image-to-3D generation tasks. We visualize the input image (\textit{column 1}) and generated RGB and normal images from novel views (\textit{column 2-7}).}
    \label{fig:app}
\end{figure}

% \begin{figure}[t]
%     \centering
%     \includegraphics[width=\linewidth]{figs/geo_refine.pdf}
    
%      NeRF + MC  \hspace{.4cm} 
%      $w/$ Geo refine   \hspace{.4cm}
%      GT \hspace{1.2cm}
%      NeRF + MC  \hspace{.4cm} 
%      $w/$ Geo refine  \hspace{.4cm}
%      GT \hspace{.6cm}
%     \caption{\textbf{Ablation study of the geometry refinement stage using mesh rendering.} We visualize the NeRF+MC results (\textit{column 1, 4}) or with (\textit{column 2, 5}) geometry refinement, and corresponding ground truth images (\textit{column 3, 6}).}
%     \label{fig:geo_refine}
% \end{figure}
\vspace{-.3cm}
\begin{figure}[t]
    \begin{minipage}{0.5\textwidth}
        \centering
        \includegraphics[width=\linewidth]{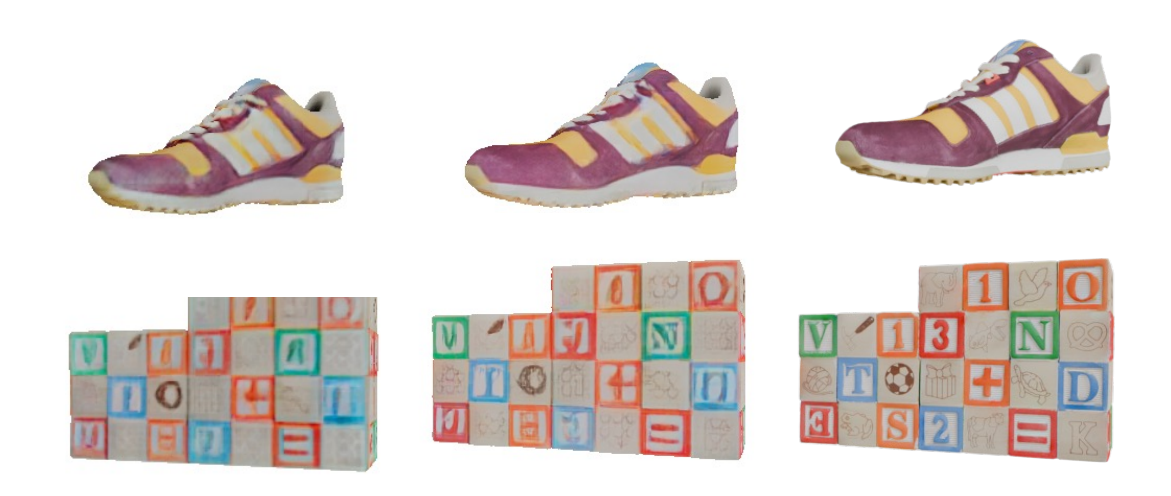}
        NeRF + MC \hspace{.2cm} NeRF+ Geo. \hspace{.8cm}  GT \hspace{.6cm}
         % Replace with your image file
        % \caption{Visual Comparison of NeRF+MC and NeRF + Mesh refinement and the ground  truth image. }

    \end{minipage}
    \hfill
    \begin{minipage}{0.45\textwidth}
        \centering
        \small
    % \begin{tabular}[t]
        % \centering
        % \caption{\textbf{Quantitative comparison on OmniObject3D dataset.}}
        %    \label{tab:omni}
        % \captionof{table}{Evaluation scores using NeRF+MC}
        \setlength{\tabcolsep}{3pt}
        \resizebox{\linewidth}{!}{%        
        \begin{tabular}{  c c c | c c}
        \midrule
          PSNR ($\uparrow$) & SSIM ($\uparrow$) & LPIPS ($\downarrow$) & CD ($\downarrow$) & IoU ($\uparrow$) \\ \midrule
        19.810 &0.903 &0.116 & 0.775 & 0.652 \\
            \midrule
        \end{tabular}
        
        }  
        \label{tab:3}
    \end{minipage}
    \caption{\textbf{Ablation study of geometry refinement}. On the left, we visualize the comparison between NeRF+MC, NeRF+Geometry refinement, and the ground-truth images. On the right, we present the evaluation scores using NeRF+MC.}
    \label{fig:geo_refine}
\end{figure}

\begin{figure}
\includegraphics[width=\linewidth]{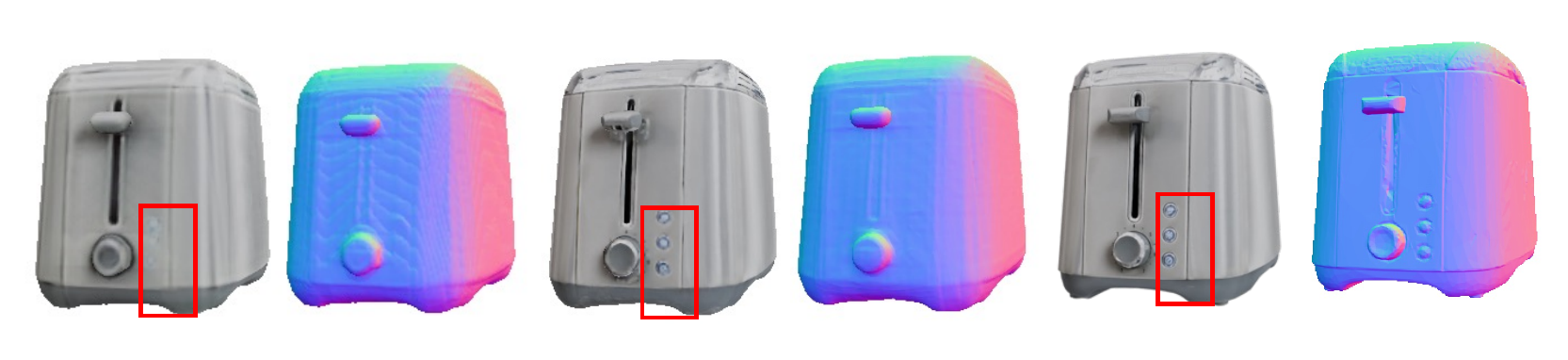}
    \centering
    \hspace{-1cm} NeRF+MC 
    \hspace{3.3cm} Ours 
    \hspace{3.5cm}  GT  \hspace{1cm}

    % \hspace{.2cm}
    % NeRF+MC \hspace{.8cm} Ours \hspace{1cm}  GT
    % \hspace{1.5cm}
    % NeRF+MC \hspace{1.cm} Ours \hspace{1.7cm}  GT\hspace{1cm}
\caption{\textbf{Additional visual comparison with Marching Cube (MC)}. When the rendering results using MC lack details (marked in red rectangles) and grid artifacts exist on surfaces. In contrast, our geometry refinement achieves better mesh rendering quality and geometry quality than the meshes extracted from NeRFs.}

    \label{fig:nerf_mc}
\end{figure}

\vspace{-.2cm}
\subsection{Ablation study}
\vspace{-.2cm}
\label{sec:43}

\begin{figure}[t]
    \centering
    \vspace{-.3cm}
    \begin{minipage}{\textwidth}
        \centering
\includegraphics[width=\linewidth]{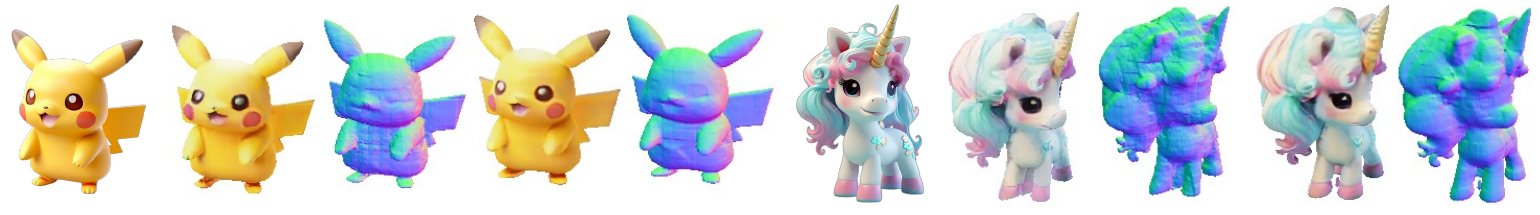}
        \hspace{.3cm}
        Input  \hspace{.8cm} $w/o\ \mathcal{L}_\text{normal}$ \hspace{.8cm} $w/\ \mathcal{L}_\text{normal}$ 
        \hspace{1.3cm}
        Input  \hspace{.8cm} $w/o \ \mathcal{L}_\text{normal}$  \hspace{.8cm} $w/\ \mathcal{L}_\text{normal}$ \hspace{.5cm}
        \caption{\textbf{Ablation on the normal loss.} The visual results show that using the normal loss could produce higher-quality surfaces.}
        \label{fig:ablate_normal}
    \end{minipage}
    
\end{figure}
\begin{figure}[t]
    \centering
    \vspace{-.3cm}
    \includegraphics[width=\linewidth]{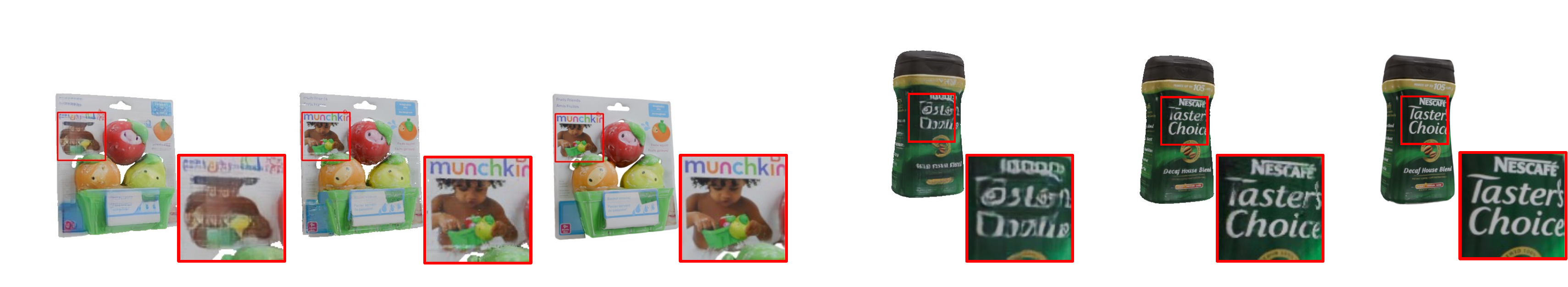}
    \vspace{-.3cm}
     \hspace{.3cm} $w/o$ Tex. refine  \hspace{.4cm} 
     $w/$ Tex. refine   \hspace{.4cm}
     GT \hspace{1cm}
     $w/o$ Tex. refine   \hspace{.4cm}
     $w/$ Tex. refine  \hspace{.4cm}
     GT 
    \caption{\textbf{Ablation study of the texture refinement procedure.} We visualize the mesh rendering results without (\textit{column 1, 4}) or with (\textit{column 2, 5}) the texture refinement procedure, and corresponding ground-truth images (\textit{column 3, 6}).}
    \vspace{-3mm}
    \label{fig:tex_refine}
\end{figure}

% \begin{figure}[t]
%     \centering
%     \includegraphics[width=\linewidth]{figs/tex_refine.pdf}
    
%      \hspace{.3cm} $w/o$ Tex. refine  \hspace{.4cm} 
%      $w/$ Tex. refine   \hspace{.4cm}
%      GT \hspace{.7cm}
%      $w/o$ Tex. refine   \hspace{.4cm}
%      $w/$ Tex. refine  \hspace{.4cm}
%      GT 
%     \caption{\textbf{Ablation study of the texture refinement procedure.} We visualize the mesh rendering results without (\textit{column 1, 4}) or with (\textit{column 2, 5}) the texture refinement procedure, and corresponding ground-truth images (\textit{column 3, 6}).}
%     \vspace{-3mm}
%     \label{fig:tex_refine}
% \end{figure}
\textbf{Geometry refinement}. 
In Fig.~\ref{fig:geo_refine}-\ref{fig:nerf_mc}, we compare results generated using NeRF+MC and NeRF with geometry refinement. We observe that directly extracting meshes from the NeRF field using MC leads to blurry texture results and a significant drop on 2D evaluation scores. In contrast, after fine-tuning at the geometry refinement stage, the rendered images show improved high-frequency details. The results in Fig.~\ref{fig:ablate_normal} present that the normal loss significantly improves the surface quality.

\textbf{Texture refinement}. We visualize the novel view mesh-rendering results generated without and with per-instance texture refinement in Fig.~\ref{fig:tex_refine}. The results of the texture refinement appear to have superior detailed textures on mesh surfaces. 

\vspace{-.2cm}

% \begin{figure}    \centering
% \includegraphics[width=\linewidth]{ICLR 2025 Template/figs/meshlrm2.pdf}
%    \hspace{2cm} Input\hspace{.3cm} Ours 
%     \hspace{.3cm} MeshLRM 
%     \hspace{.4cm} Input 
%     \hspace{.3cm} Ours 
%     \hspace{.3cm} MeshLRM 
%     \hspace{.3cm} 
%     Input\hspace{.3cm} Ours 
%     \hspace{.3cm} MeshLRM 
%     \caption{\textbf{Visual comparison with MeshLRM}~\citep{meshlrm} on single image input.}
%     \label{fig:meshlrm2}
% \end{figure}

% \begin{figure}
%     \centering
%     \includegraphics[width=\linewidth]{ICLR 2025 Template/figs/meshlrm.pdf}
%     \centering
%     $\qquad$ Input \hspace{1.5cm} Ours \hspace{1.5cm} MeshLRM
%     \hspace{1cm}
%     Input
%     \hspace{1cm} Ours \hspace{.9cm} MeshLRM
%     \caption{\textbf{Visual comparison with MeshLRM}~\citep{meshlrm} on synthetic input images. Our results are comparable to those of the concurrent MeshLRM~\citep{meshlrm}.}
%     \label{fig:meshlrm}
% \end{figure}
% \input{ICLR 2025 Template/figs_latex/fig_meshlrm}

\vspace{-.2cm}
\section{Conclusion}
\label{sec:conclusion}
\vspace{-.3cm}
In this work, we introduce GTR, a large 3D reconstruction model that takes multi-view images as input. Our approach enables the generation of high-quality meshes with faithful texture reconstruction within seconds. We achieve this through three key contributions: modifications to the current LRM model architecture, the integration of end-to-end geometry refinement with NeRF initialization, and the implementation of a per-instance texture refinement procedure.
% Extensive experiments and evaluations conducted in both 2D and 3D spaces demonstrate the state-of-the-art performance of our approach. Additionally, our approach can be applied to various downstream applications, such as text/image-to-3D generation.

\clearpage
\bibliography{iclr2025_conference}

\begin{thebibliography}{46}
\providecommand{\natexlab}[1]{#1}
\providecommand{\url}[1]{\texttt{#1}}
\expandafter\ifx\csname urlstyle\endcsname\relax
  \providecommand{\doi}[1]{doi: #1}\else
  \providecommand{\doi}{doi: \begingroup \urlstyle{rm}\Url}\fi

\bibitem[Blattmann et~al.(2023)Blattmann, Dockhorn, Kulal, Mendelevitch, Kilian, Lorenz, Levi, English, Voleti, Letts, et~al.]{svd}
Andreas Blattmann, Tim Dockhorn, Sumith Kulal, Daniel Mendelevitch, Maciej Kilian, Dominik Lorenz, Yam Levi, Zion English, Vikram Voleti, Adam Letts, et~al.
\newblock Stable video diffusion: Scaling latent video diffusion models to large datasets.
\newblock \emph{arXiv preprint arXiv:2311.15127}, 2023.

\bibitem[Campbell et~al.(2008)Campbell, Vogiatzis, Hern{\'a}ndez, and Cipolla]{campbell2008using}
Neill~DF Campbell, George Vogiatzis, Carlos Hern{\'a}ndez, and Roberto Cipolla.
\newblock Using multiple hypotheses to improve depth-maps for multi-view stereo.
\newblock In \emph{ECCV}, 2008.

\bibitem[Caron et~al.(2021)Caron, Touvron, Misra, J{\'e}gou, Mairal, Bojanowski, and Joulin]{DINO}
Mathilde Caron, Hugo Touvron, Ishan Misra, Herv{\'e} J{\'e}gou, Julien Mairal, Piotr Bojanowski, and Armand Joulin.
\newblock Emerging properties in self-supervised vision transformers.
\newblock In \emph{ICCV}, 2021.

\bibitem[Chen et~al.(2023)Chen, Chen, Jiao, and Jia]{fantasia3d}
Rui Chen, Yongwei Chen, Ningxin Jiao, and Kui Jia.
\newblock Fantasia3d: Disentangling geometry and appearance for high-quality text-to-3d content creation.
\newblock In \emph{ICCV}, 2023.

\bibitem[Deitke et~al.(2023)Deitke, Schwenk, Salvador, Weihs, Michel, VanderBilt, Schmidt, Ehsani, Kembhavi, and Farhadi]{objaverse}
Matt Deitke, Dustin Schwenk, Jordi Salvador, Luca Weihs, Oscar Michel, Eli VanderBilt, Ludwig Schmidt, Kiana Ehsani, Aniruddha Kembhavi, and Ali Farhadi.
\newblock Objaverse: A universe of annotated 3d objects.
\newblock In \emph{CVPR}, 2023.

\bibitem[Downs et~al.(2022)Downs, Francis, Koenig, Kinman, Hickman, Reymann, McHugh, and Vanhoucke]{gso}
Laura Downs, Anthony Francis, Nate Koenig, Brandon Kinman, Ryan Hickman, Krista Reymann, Thomas~B McHugh, and Vincent Vanhoucke.
\newblock Google scanned objects: A high-quality dataset of 3d scanned household items.
\newblock In \emph{ICRA}, 2022.

\bibitem[Gallup et~al.(2007)Gallup, Frahm, Mordohai, Yang, and Pollefeys]{gallup2007real}
David Gallup, Jan-Michael Frahm, Philippos Mordohai, Qingxiong Yang, and Marc Pollefeys.
\newblock Real-time plane-sweeping stereo with multiple sweeping directions.
\newblock In \emph{CVPR}, 2007.

\bibitem[Gu et~al.(2020)Gu, Fan, Zhu, Dai, Tan, and Tan]{gu2020cascade}
Xiaodong Gu, Zhiwen Fan, Siyu Zhu, Zuozhuo Dai, Feitong Tan, and Ping Tan.
\newblock Cascade cost volume for high-resolution multi-view stereo and stereo matching.
\newblock In \emph{CVPR}, 2020.

\bibitem[Hong et~al.(2024)Hong, Zhang, Gu, Bi, Zhou, Liu, Liu, Sunkavalli, Bui, and Tan]{lrm}
Yicong Hong, Kai Zhang, Jiuxiang Gu, Sai Bi, Yang Zhou, Difan Liu, Feng Liu, Kalyan Sunkavalli, Trung Bui, and Hao Tan.
\newblock Lrm: Large reconstruction model for single image to 3d.
\newblock In \emph{ICLR}, 2024.

\bibitem[Kutulakos \& Seitz(2000)Kutulakos and Seitz]{kutulakos2000theory}
Kiriakos~N Kutulakos and Steven~M Seitz.
\newblock A theory of shape by space carving.
\newblock \emph{IJCV}, 2000.

\bibitem[Lhuillier \& Quan(2005)Lhuillier and Quan]{lhuillier2005quasi}
Maxime Lhuillier and Long Quan.
\newblock A quasi-dense approach to surface reconstruction from uncalibrated images.
\newblock \emph{TPAMI}, 2005.

\bibitem[Li et~al.(2024)Li, Tan, Zhang, Xu, Luan, Xu, Hong, Sunkavalli, Shakhnarovich, and Bi]{instant3d}
Jiahao Li, Hao Tan, Kai Zhang, Zexiang Xu, Fujun Luan, Yinghao Xu, Yicong Hong, Kalyan Sunkavalli, Greg Shakhnarovich, and Sai Bi.
\newblock Instant3d: Fast text-to-3d with sparse-view generation and large reconstruction model.
\newblock In \emph{ICLR}, 2024.

\bibitem[Lin et~al.(2023{\natexlab{a}})Lin, Gao, Tang, Takikawa, Zeng, Huang, Kreis, Fidler, Liu, and Lin]{magic3d}
Chen-Hsuan Lin, Jun Gao, Luming Tang, Towaki Takikawa, Xiaohui Zeng, Xun Huang, Karsten Kreis, Sanja Fidler, Ming-Yu Liu, and Tsung-Yi Lin.
\newblock Magic3d: High-resolution text-to-3d content creation.
\newblock In \emph{CVPR}, 2023{\natexlab{a}}.

\bibitem[Lin et~al.(2023{\natexlab{b}})Lin, Lee, Menapace, Chai, Siarohin, Yang, and Tulyakov]{inifinicity}
Chieh~Hubert Lin, Hsin-Ying Lee, Willi Menapace, Menglei Chai, Aliaksandr Siarohin, Ming-Hsuan Yang, and Sergey Tulyakov.
\newblock Infinicity: Infinite-scale city synthesis.
\newblock In \emph{ICCV}, 2023{\natexlab{b}}.

\bibitem[Liu et~al.(2023)Liu, Wu, Van~Hoorick, Tokmakov, Zakharov, and Vondrick]{zero123}
Ruoshi Liu, Rundi Wu, Basile Van~Hoorick, Pavel Tokmakov, Sergey Zakharov, and Carl Vondrick.
\newblock Zero-1-to-3: Zero-shot one image to 3d object.
\newblock In \emph{ICCV}, 2023.

\bibitem[Liu et~al.(2024)Liu, Lin, Zeng, Long, Liu, Komura, and Wang]{syncdreamer}
Yuan Liu, Cheng Lin, Zijiao Zeng, Xiaoxiao Long, Lingjie Liu, Taku Komura, and Wenping Wang.
\newblock Syncdreamer: Generating multiview-consistent images from a single-view image.
\newblock In \emph{ICLR}, 2024.

\bibitem[Long et~al.(2023)Long, Guo, Lin, Liu, Dou, Liu, Ma, Zhang, Habermann, Theobalt, et~al.]{wonder3d}
Xiaoxiao Long, Yuan-Chen Guo, Cheng Lin, Yuan Liu, Zhiyang Dou, Lingjie Liu, Yuexin Ma, Song-Hai Zhang, Marc Habermann, Christian Theobalt, et~al.
\newblock Wonder3d: Single image to 3d using cross-domain diffusion.
\newblock In \emph{CVPR}, 2023.

\bibitem[Lorensen \& Cline(1998)Lorensen and Cline]{mc}
William~E Lorensen and Harvey~E Cline.
\newblock Marching cubes: A high resolution 3d surface construction algorithm.
\newblock In \emph{Seminal graphics: pioneering efforts that shaped the field}, pp.\  347--353. 1998.

\bibitem[Ma et~al.(2021)Ma, Gong, Wang, Huang, Chen, and Yu]{ma2021epp}
Xinjun Ma, Yue Gong, Qirui Wang, Jingwei Huang, Lei Chen, and Fan Yu.
\newblock Epp-mvsnet: Epipolar-assembling based depth prediction for multi-view stereo.
\newblock In \emph{ICCV}, 2021.

\bibitem[Odena et~al.(2016)Odena, Dumoulin, and Olah]{deconv}
Augustus Odena, Vincent Dumoulin, and Chris Olah.
\newblock Deconvolution and checkerboard artifacts.
\newblock \emph{Distill}, 2016.
\newblock \doi{10.23915/distill.00003}.
\newblock URL \url{http://distill.pub/2016/deconv-checkerboard}.

\bibitem[Poole et~al.(2023)Poole, Jain, Barron, and Mildenhall]{dreamfusion}
Ben Poole, Ajay Jain, Jonathan~T Barron, and Ben Mildenhall.
\newblock Dreamfusion: Text-to-3d using 2d diffusion.
\newblock In \emph{ICLR}, 2023.

\bibitem[Qian et~al.(2024)Qian, Mai, Hamdi, Ren, Siarohin, Li, Lee, Skorokhodov, Wonka, Tulyakov, et~al.]{magic123}
Guocheng Qian, Jinjie Mai, Abdullah Hamdi, Jian Ren, Aliaksandr Siarohin, Bing Li, Hsin-Ying Lee, Ivan Skorokhodov, Peter Wonka, Sergey Tulyakov, et~al.
\newblock Magic123: One image to high-quality 3d object generation using both 2d and 3d diffusion priors.
\newblock In \emph{ICLR}, 2024.

\bibitem[Rombach et~al.(2022)Rombach, Blattmann, Lorenz, Esser, and Ommer]{ldm}
Robin Rombach, Andreas Blattmann, Dominik Lorenz, Patrick Esser, and Bj{\"o}rn Ommer.
\newblock High-resolution image synthesis with latent diffusion models.
\newblock In \emph{CVPR}, 2022.

\bibitem[Saharia et~al.(2022)Saharia, Chan, Saxena, Li, Whang, Denton, Ghasemipour, Gontijo~Lopes, Karagol~Ayan, Salimans, et~al.]{imagen}
Chitwan Saharia, William Chan, Saurabh Saxena, Lala Li, Jay Whang, Emily~L Denton, Kamyar Ghasemipour, Raphael Gontijo~Lopes, Burcu Karagol~Ayan, Tim Salimans, et~al.
\newblock Photorealistic text-to-image diffusion models with deep language understanding.
\newblock In \emph{NeurIPS}, 2022.

\bibitem[Seitz \& Dyer(1999)Seitz and Dyer]{seitz1999photorealistic}
Steven~M Seitz and Charles~R Dyer.
\newblock Photorealistic scene reconstruction by voxel coloring.
\newblock \emph{IJCV}, 1999.

\bibitem[Shen et~al.(2021)Shen, Gao, Yin, Liu, and Fidler]{dmtet}
Tianchang Shen, Jun Gao, Kangxue Yin, Ming-Yu Liu, and Sanja Fidler.
\newblock Deep marching tetrahedra: a hybrid representation for high-resolution 3d shape synthesis.
\newblock In \emph{NeurIPS}, 2021.

\bibitem[Shen et~al.(2023)Shen, Munkberg, Hasselgren, Yin, Wang, Chen, Gojcic, Fidler, Sharp, and Gao]{flexicube}
Tianchang Shen, Jacob Munkberg, Jon Hasselgren, Kangxue Yin, Zian Wang, Wenzheng Chen, Zan Gojcic, Sanja Fidler, Nicholas Sharp, and Jun Gao.
\newblock Flexible isosurface extraction for gradient-based mesh optimization.
\newblock \emph{TOG}, 2023.

\bibitem[Shi et~al.(2023)Shi, Chen, Zhang, Liu, Xu, Wei, Chen, Zeng, and Su]{zero123pp}
Ruoxi Shi, Hansheng Chen, Zhuoyang Zhang, Minghua Liu, Chao Xu, Xinyue Wei, Linghao Chen, Chong Zeng, and Hao Su.
\newblock Zero123++: a single image to consistent multi-view diffusion base model.
\newblock In \emph{arXiv preprint arXiv:2310.15110}, 2023.

\bibitem[Shi et~al.(2016)Shi, Caballero, Husz{\'a}r, Totz, Aitken, Bishop, Rueckert, and Wang]{pixelshuffle}
Wenzhe Shi, Jose Caballero, Ferenc Husz{\'a}r, Johannes Totz, Andrew~P Aitken, Rob Bishop, Daniel Rueckert, and Zehan Wang.
\newblock Real-time single image and video super-resolution using an efficient sub-pixel convolutional neural network.
\newblock In \emph{CVPR}, 2016.

\bibitem[Shi et~al.(2024)Shi, Wang, Ye, Long, Li, and Yang]{mvdream}
Yichun Shi, Peng Wang, Jianglong Ye, Mai Long, Kejie Li, and Xiao Yang.
\newblock Mvdream: Multi-view diffusion for 3d generation.
\newblock In \emph{ICLR}, 2024.

\bibitem[Stereopsis(2010)]{stereopsis2010accurate}
Robust~Multiview Stereopsis.
\newblock Accurate, dense, and robust multiview stereopsis.
\newblock \emph{TPAMI}, 2010.

\bibitem[Tang et~al.(2024)Tang, Chen, Chen, Wang, Zeng, and Liu]{lgm}
Jiaxiang Tang, Zhaoxi Chen, Xiaokang Chen, Tengfei Wang, Gang Zeng, and Ziwei Liu.
\newblock Lgm: Large multi-view gaussian model for high-resolution 3d content creation.
\newblock \emph{arXiv preprint arXiv:2402.05054}, 2024.

\bibitem[Voleti et~al.(2024)Voleti, Yao, Boss, Letts, Pankratz, Tochilkin, Laforte, Rombach, and Jampani]{sv3d}
Vikram Voleti, Chun-Han Yao, Mark Boss, Adam Letts, David Pankratz, Dmitry Tochilkin, Christian Laforte, Robin Rombach, and Varun Jampani.
\newblock Sv3d: Novel multi-view synthesis and 3d generation from a single image using latent video diffusion.
\newblock \emph{arXiv preprint arXiv:2403.12008}, 2024.

\bibitem[Wang et~al.(2021)Wang, Galliani, Vogel, Speciale, and Pollefeys]{wang2021patchmatchnet}
Fangjinhua Wang, Silvano Galliani, Christoph Vogel, Pablo Speciale, and Marc Pollefeys.
\newblock Patchmatchnet: Learned multi-view patchmatch stereo.
\newblock In \emph{CVPR}, 2021.

\bibitem[Wang et~al.(2023{\natexlab{a}})Wang, Du, Li, Yeh, and Shakhnarovich]{sjc}
Haochen Wang, Xiaodan Du, Jiahao Li, Raymond~A Yeh, and Greg Shakhnarovich.
\newblock Score jacobian chaining: Lifting pretrained 2d diffusion models for 3d generation.
\newblock In \emph{CVPR}, 2023{\natexlab{a}}.

\bibitem[Wang et~al.(2023{\natexlab{b}})Wang, Lu, Wang, Bao, Li, Su, and Zhu]{prolificdreamer}
Zhengyi Wang, Cheng Lu, Yikai Wang, Fan Bao, Chongxuan Li, Hang Su, and Jun Zhu.
\newblock Prolificdreamer: High-fidelity and diverse text-to-3d generation with variational score distillation.
\newblock In \emph{NeurIPS}, 2023{\natexlab{b}}.

\bibitem[Wang et~al.(2024)Wang, Wang, Chen, Xiang, Chen, Yu, Li, Su, and Zhu]{crm}
Zhengyi Wang, Yikai Wang, Yifei Chen, Chendong Xiang, Shuo Chen, Dajiang Yu, Chongxuan Li, Hang Su, and Jun Zhu.
\newblock Crm: Single image to 3d textured mesh with convolutional reconstruction model.
\newblock \emph{arXiv preprint arXiv:2403.05034}, 2024.

\bibitem[Wei et~al.(2023)Wei, Xiang, Bi, Chen, Sunkavalli, Xu, and Su]{diffmc}
Xinyue Wei, Fanbo Xiang, Sai Bi, Anpei Chen, Kalyan Sunkavalli, Zexiang Xu, and Hao Su.
\newblock Neumanifold: Neural watertight manifold reconstruction with efficient and high-quality rendering support.
\newblock \emph{arXiv preprint arXiv:2305.17134}, 2023.

\bibitem[Wei et~al.(2024)Wei, Zhang, Bi, Tan, Luan, Deschaintre, Sunkavalli, Su, and Xu]{meshlrm}
Xinyue Wei, Kai Zhang, Sai Bi, Hao Tan, Fujun Luan, Valentin Deschaintre, Kalyan Sunkavalli, Hao Su, and Zexiang Xu.
\newblock Meshlrm: Large reconstruction model for high-quality mesh.
\newblock \emph{arXiv preprint arXiv:2404.12385}, 2024.

\bibitem[Wei et~al.(2021)Wei, Zhu, Min, Chen, and Wang]{wei2021aa}
Zizhuang Wei, Qingtian Zhu, Chen Min, Yisong Chen, and Guoping Wang.
\newblock Aa-rmvsnet: Adaptive aggregation recurrent multi-view stereo network.
\newblock In \emph{ICCV}, 2021.

\bibitem[Wu et~al.(2023)Wu, Zhang, Fu, Wang, Ren, Pan, Wu, Yang, Wang, Qian, et~al.]{omniobject3d}
Tong Wu, Jiarui Zhang, Xiao Fu, Yuxin Wang, Jiawei Ren, Liang Pan, Wayne Wu, Lei Yang, Jiaqi Wang, Chen Qian, et~al.
\newblock Omniobject3d: Large-vocabulary 3d object dataset for realistic perception, reconstruction and generation.
\newblock In \emph{CVPR}, 2023.

\bibitem[Xu et~al.(2024{\natexlab{a}})Xu, Cheng, Gao, Wang, Gao, and Shan]{instantmesh}
Jiale Xu, Weihao Cheng, Yiming Gao, Xintao Wang, Shenghua Gao, and Ying Shan.
\newblock Instantmesh: Efficient 3d mesh generation from a single image with sparse-view large reconstruction models.
\newblock \emph{arXiv preprint arXiv:2404.07191}, 2024{\natexlab{a}}.

\bibitem[Xu et~al.(2024{\natexlab{b}})Xu, Shi, Yifan, Chen, Yang, Peng, Shen, and Wetzstein]{grm}
Yinghao Xu, Zifan Shi, Wang Yifan, Hansheng Chen, Ceyuan Yang, Sida Peng, Yujun Shen, and Gordon Wetzstein.
\newblock Grm: Large gaussian reconstruction model for efficient 3d reconstruction and generation.
\newblock \emph{arXiv preprint arXiv:2403.14621}, 2024{\natexlab{b}}.

\bibitem[Xu et~al.(2024{\natexlab{c}})Xu, Tan, Luan, Bi, Wang, Li, Shi, Sunkavalli, Wetzstein, Xu, et~al.]{dmv3d}
Yinghao Xu, Hao Tan, Fujun Luan, Sai Bi, Peng Wang, Jiahao Li, Zifan Shi, Kalyan Sunkavalli, Gordon Wetzstein, Zexiang Xu, et~al.
\newblock Dmv3d: Denoising multi-view diffusion using 3d large reconstruction model.
\newblock In \emph{ICLR}, 2024{\natexlab{c}}.

\bibitem[Yi et~al.(2020)Yi, Wei, Ding, Zhang, Chen, Wang, and Tai]{yi2020pyramid}
Hongwei Yi, Zizhuang Wei, Mingyu Ding, Runze Zhang, Yisong Chen, Guoping Wang, and Yu-Wing Tai.
\newblock Pyramid multi-view stereo net with self-adaptive view aggregation.
\newblock In \emph{ECCV}, 2020.

\bibitem[Zhu \& Zhuang(2023)Zhu and Zhuang]{hifa}
Joseph Zhu and Peiye Zhuang.
\newblock Hifa: High-fidelity text-to-3d with advanced diffusion guidance.
\newblock In \emph{ICLR}, 2023.

\end{thebibliography}
\bibliographystyle{iclr2025_conference}

\clearpage
\appendix

We present additional ablation study, limitation, and visual results in the \textbf{Appendix manuscript}. Please also refer to our \textbf{website demo} in the supplementary material for a comprehensive overview.

\section{Ablation study}
\label{sup:ablation}
\begin{figure}[h]
    \centering
    \begin{minipage}{0.46\linewidth}
        \includegraphics[width=\linewidth]{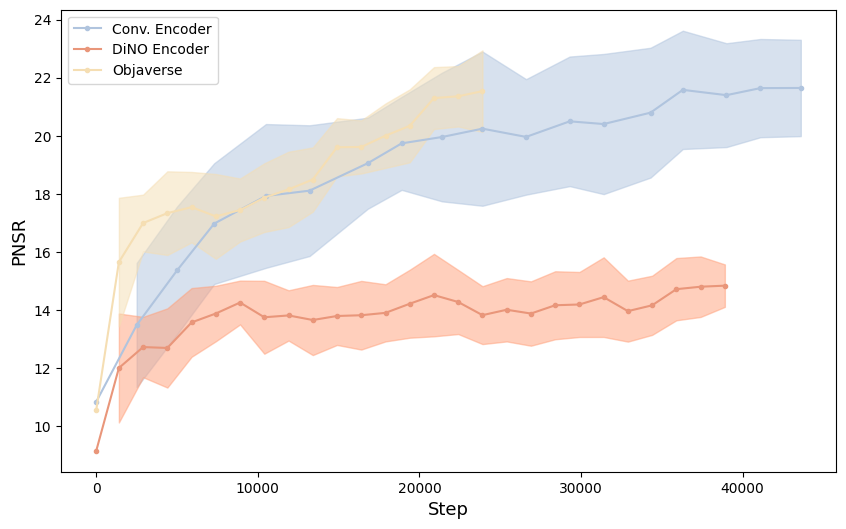}
        \caption{\textbf{Ablation study on image encoders} ( Conv. vs DiNO) and dataset. }
        \label{fig:dino}
    \end{minipage}
    \begin{minipage}{0.46\linewidth}
       \includegraphics[width=\linewidth]{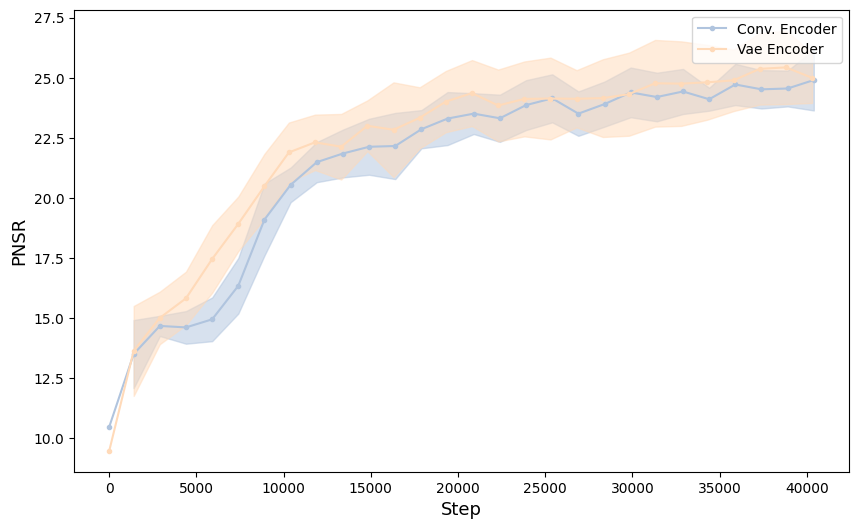}
      \caption{\textbf{Ablation study on image encoders} (Conv. vs VAE).}
      \label{fig:vae}
    \end{minipage}
    
\end{figure}

\textbf{DiNO encoder}. We conduct experiments using different encoders. In Fig.~\ref{fig:dino} we present the validation PSNR curves during training, when using either our convolutional image encoder or a pre-trained DiNO ViT~\citep{DINO}. Specifically, our convolutional encoder is a single layer that downsamples input images from 512 to 32. The triplane generator is a self-attention transformer, identical in both settings. We train both models on 8 80G A100 GPUs. We observe that the DiNO experiment did not show improved convergence during the initial iterations. Alternatively, more careful designs could optimize the use of DiNO ViT, which we leave for future study.

\textbf{Objaverse training dataset}. In Fig.~\ref{fig:dino}, we also show the training process with a dataset consisting solely of 100k Objaverse~\citep{objaverse} images. We did not observe a performance drop in the early stage compared to the other experiments in the figure, which were trained on our mixed dataset.

\textbf{Vae encoder}.  In Fig.\ref{fig:vae}, we show preliminary results using a pretrained VAE encoder\footnote{In practice, we use the pretrained SD VAE from https://huggingface.co/madebyollin/taesd} from an SD model~\citep{ldm}. To enable the VAE encoder to handle multi-channel input, we separately provide images, masks, and the camera rays to the encoder, then assemble the output features using a convolution layer. Experiments are run on 32 80G A100 GPUs. We observe that using a pretrained VAE encoder leads to better convergence in the early training stage. We attribute this to the good initialization provided by VAEs compared to training the convolutional encoder from scratch.

\section{Limitation} 

Current triplane features at a resolution of $256$ are sometimes insufficient to capture fine geometric details. Meanwhile, alternative representations — explicit (e.g., sparse voxels) or implicit (e.g., vector sets) — offer promising avenues for exploration. Additionally, our approach currently requires camera-conditioned input. It would be more advantageous to develop methods capable of handling unposed input. Moreover, our work focuses on object-centric assets, leaving the exploration of complex scenes with composite objects as a valuable direction for future research.

\section{Additional results}

\label{sup:visual}
\begin{figure}
    \centering
    \includegraphics[width=\linewidth]{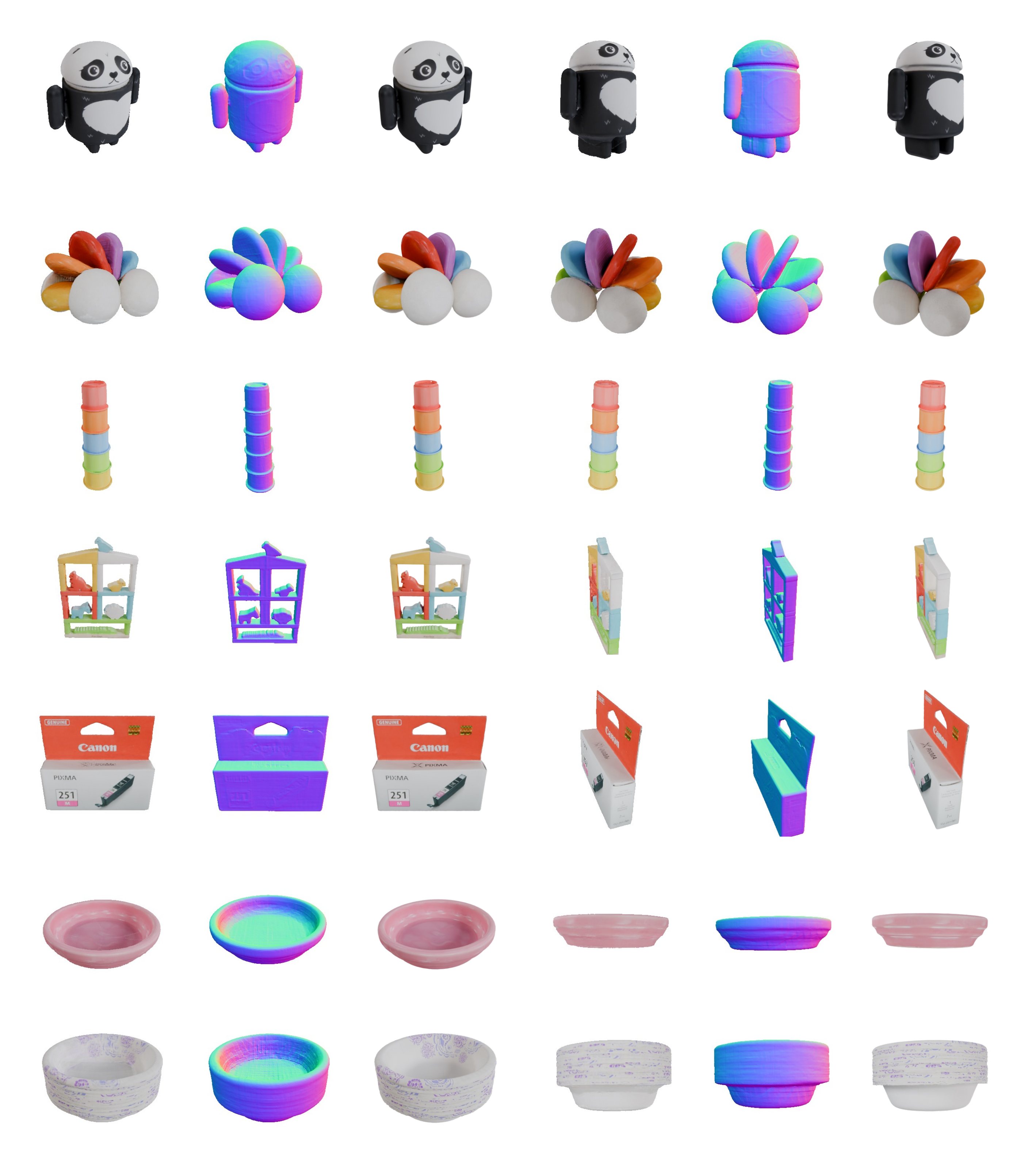}
    \caption{\textbf{Additional visual 3D reconstruction results on GSO~\citep{gso} dataset}. The input of our model is 4 orthogonal views. We show the novel view generated RGB images (\textit{column 1, 4}) and normal images (\textit{column 2, 5}), and the ground-truth images (\textit{column 3, 6}).}
    \label{fig:sup_gso1}
\end{figure}

\begin{figure}
    \centering
    \includegraphics[width=\linewidth]{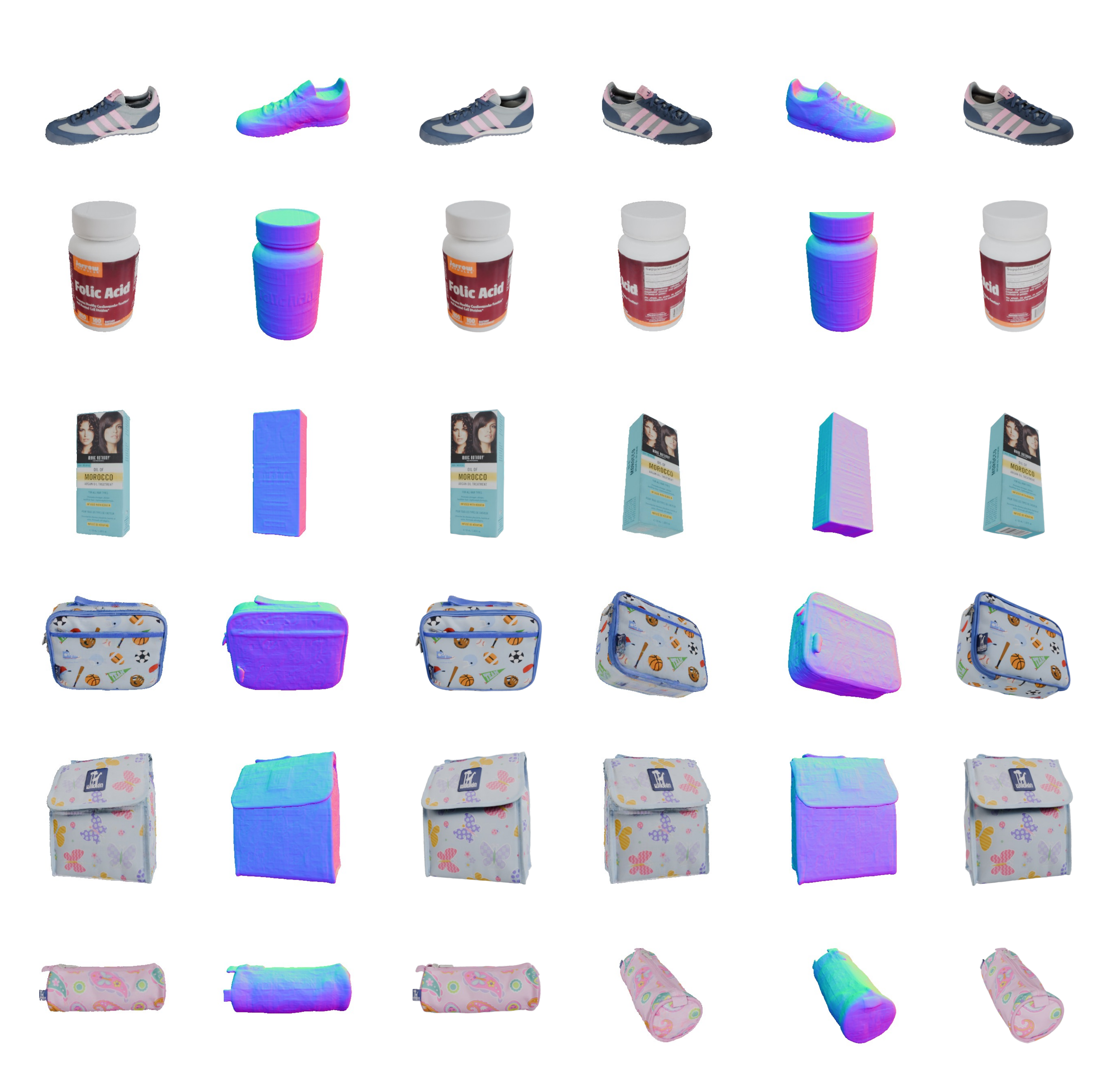}
    \caption{\textbf{Additional visual 3D reconstruction results on GSO~\citep{gso} dataset}. The input of our model is 4 orthogonal views. We show the novel view generated RGB images (\textit{column 1, 4}) and normal images (\textit{column 2, 5}), and the ground-truth images (\textit{column 3, 6}).}
    \label{fig:sup_gso2}
\end{figure}

\begin{figure}
    \centering
    \includegraphics[width=\linewidth]{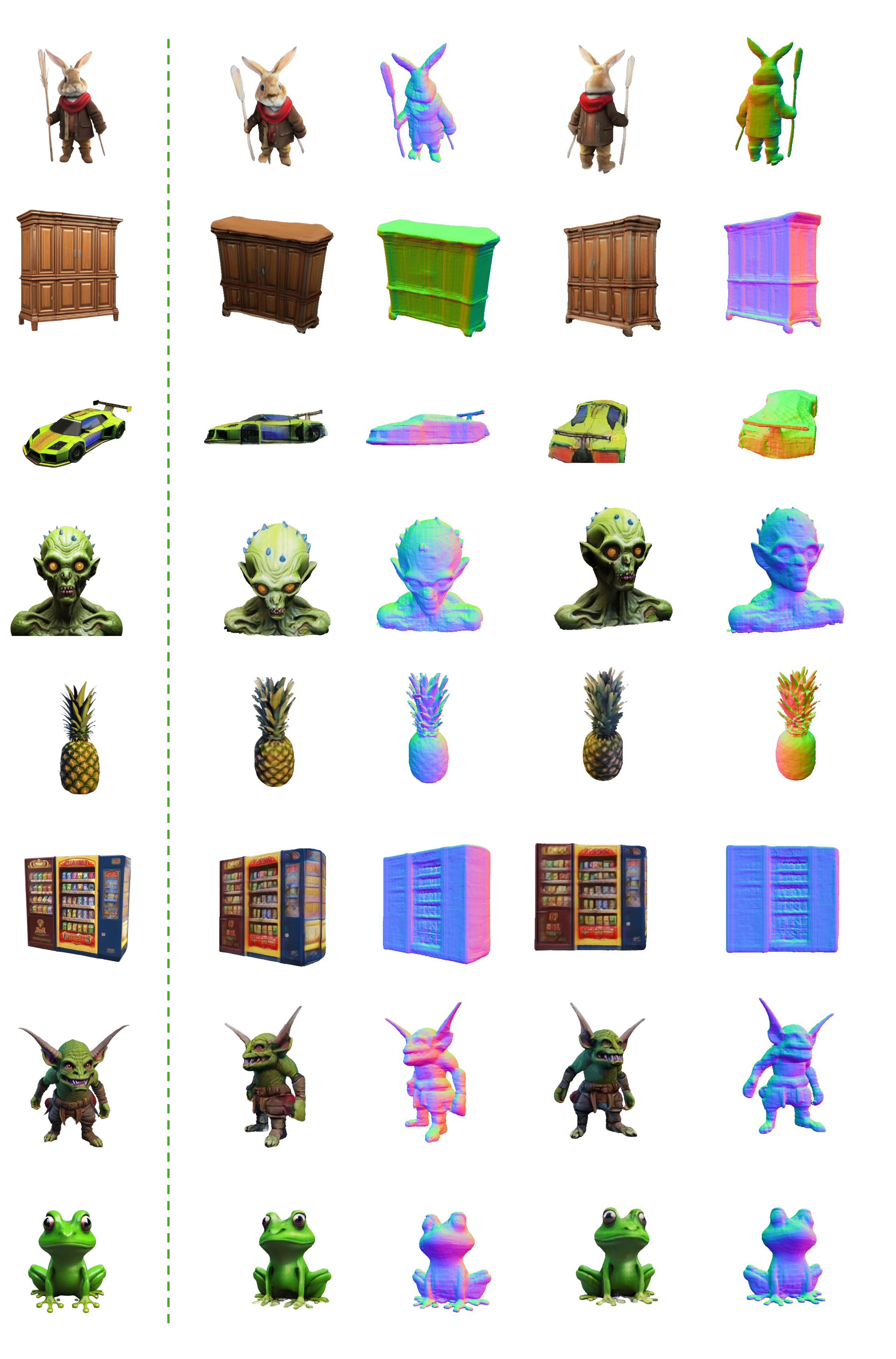}
    \caption{\textbf{Additional visual 3D asset generation results}. The input images (\textit{column 1}) are either generated from text using pre-trained text-to-image diffusion models~\citep{ldm} or online generated images.}
    \label{fig:sup_gen1}
\end{figure}

We show additional results generated by our approach in Fig.~\ref{fig:sup_gso1}-~\ref{fig:sup_gen1}.

% \begin{figure}
%     \centering
%     \includegraphics[width=\linewidth]{sup_figs/sup_gso1.pdf}
%     \caption{\textbf{Additional visual 3D reconstruction results on GSO~\citep{gso} dataset}. We show the novel view generated RGB images (\textit{column 1, 4}) and normal images (\textit{column 2, 5}), and the ground-truth images (\textit{column 3, 6}).}
%     \label{fig:sup_gso1}
% \end{figure}

% \begin{figure}
%     \centering
%     \includegraphics[width=\linewidth]{sup_figs/sup_gso1.pdf}
%     \caption{\textbf{Additional visual 3D reconstruction results on GSO~\citep{gso} dataset}. We show the novel view generated RGB images (\textit{column 1, 4}) and normal images (\textit{column 2, 5}), and the ground-truth images (\textit{column 3, 6}).}
%     \label{fig:sup_gso1}
% \end{figure}

\end{document}